\documentclass{article}

\usepackage{arxiv}

\usepackage[utf8]{inputenc} 
\usepackage[T1]{fontenc}    
\usepackage{hyperref}       
\usepackage{url}            
\usepackage{booktabs}       
\usepackage{amsfonts}       
\usepackage{nicefrac}       
\usepackage{microtype}      
\usepackage{lipsum}		
\usepackage{graphicx}
\usepackage{natbib}
\usepackage{doi}
\usepackage{listings}
\usepackage{float}
\usepackage{amsmath}
\usepackage{bbm}
\usepackage[flushleft]{threeparttable} 
\usepackage{caption}
\usepackage{array} 
\usepackage{tabularx}
\usepackage{multirow}
\usepackage[table]{xcolor}

\title{When language and vision meet road safety: leveraging multimodal large language models for video-based traffic accident analysis}

\date{} 					

\author{
    \textbf{Ruixuan Zhang} \quad
    \textbf{Beichen Wang}\quad
    \textbf{Juexiao Zhang} \quad
    \textbf{Zilin Bian}$^{*}$ \\[1ex]
    \textbf{Chen Feng}$^{\dag}$\quad
    \textbf{Kaan Ozbay}$^{\dag}$\\ \\
    New York University\\
    \texttt{\{ruixuan.zhang,bw2716,jz4725,zb536,cfeng,kaan.ozbay\}@nyu.edu}
}

\renewcommand{\shorttitle}{Multimodal Large Language Models for Video-based Traffic Accident Analysis}

\begin{document}
\maketitle

\vspace{-2em} 
\renewcommand{\thefootnote}{} 
\footnotetext{\hspace{-1em}$^{*}$Corresponding author. $^{\dag}$Equal advising.}

\begin{abstract}
The increasing availability of traffic videos functioning on a 24/7/365 time scale has the great potential of increasing the spatio-temporal coverage of traffic accidents, which will help improve traffic safety. However, analyzing footage from hundreds, if not thousands, of traffic cameras in a 24/7/365 working protocol remains an extremely challenging task, as current vision-based approaches primarily focus on extracting raw information, such as vehicle trajectories or individual object detection, but require laborious post-processing to derive actionable insights. We propose SeeUnsafe, a new framework that integrates Multimodal Large Language Model (MLLM) agents to transform video-based traffic accident analysis from a traditional extraction-then-explanation workflow to a more interactive, conversational approach. This shift significantly enhances processing throughput by automating complex tasks like video classification and visual grounding, while improving adaptability by enabling seamless adjustments to diverse traffic scenarios and user-defined queries. Our framework employs a severity-based aggregation strategy to handle videos of various lengths and a novel multimodal prompt to generate structured responses for review and evaluation and enable fine-grained visual grounding. We introduce IMS (Information Matching Score), a new MLLM-based metric for aligning structured responses with ground truth. We conduct extensive experiments on the Toyota Woven Traffic Safety dataset, demonstrating that SeeUnsafe effectively performs accident-aware video classification and visual grounding by leveraging off-the-shelf MLLMs. Source code will be available at \url{https://github.com/ai4ce/SeeUnsafe}.
\end{abstract}


\section{Introduction}
\label{sec: introduction}
The extensive deployment of traffic cameras results in the daily accumulation of vast amounts of video footage. These recordings contain crucial information regarding traffic safety, such as vehicle conflicts, pedestrian-vehicle interactions, and other risk-related events. With advancements in computer vision technologies, it is now possible to extract this valuable safety-related data, offering transportation managers insights to enhance safety and mobility performance across various neighborhoods within expansive road networks. Prior works have explored utilizing publicly available video streams from over 900 traffic cameras throughout New York City for diverse urban management studies, including social distancing analysis \cite{zuo2021reference}, illegal parking detection \cite{gao2022new}, work zone identification \cite{zuo2023urban} and mobility monitoring \cite{li2024multi}. These initial efforts underscore the potential for more informative and effective urban traffic safety management through video and image analysis.

While extensive camera coverage yields valuable traffic scene videos, thoroughly understanding these videos across spatial and temporal dimensions demands significant manual effort and traffic safety expertise. Even with automated image processing techniques like object detection and tracking, identifying recordings relevant to traffic safety analysis remains challenging due to the lack of comprehensive scene-level analysis that considers multiple contributing factors. Beyond merely extracting information from raw visual inputs, leveraging cameras within a large-scale road network for comprehensive accident analysis necessitates the integration of perception results from each camera and deriving safety insights from these distributed sources. The increasing availability of traffic video footage presents a significant opportunity for advancing traffic safety analysis. However, the rarity of safety-critical events, such as collisions and near-misses, buried within the vast volume of routine recordings poses a substantial challenge for transportation managers and researchers to work with. This disparity between the abundance of video footage and the rarity of safety-critical events highlights the need for an automated framework capable of efficiently extracting meaningful safety-related insights from massive amounts of data at scale. Without such a solution, fully utilizing traffic camera footage to enhance traffic safety remains an overwhelming and impractical task. Therefore, there is a pressing need to free experts from complex image processing tasks, allowing them to focus on minimizing life and property loss, which motivates us to find a solution that eases information extraction and supports the primary goal of improving road safety.

\begin{figure}[t]
	\centering
	\includegraphics[width=.8\textwidth]{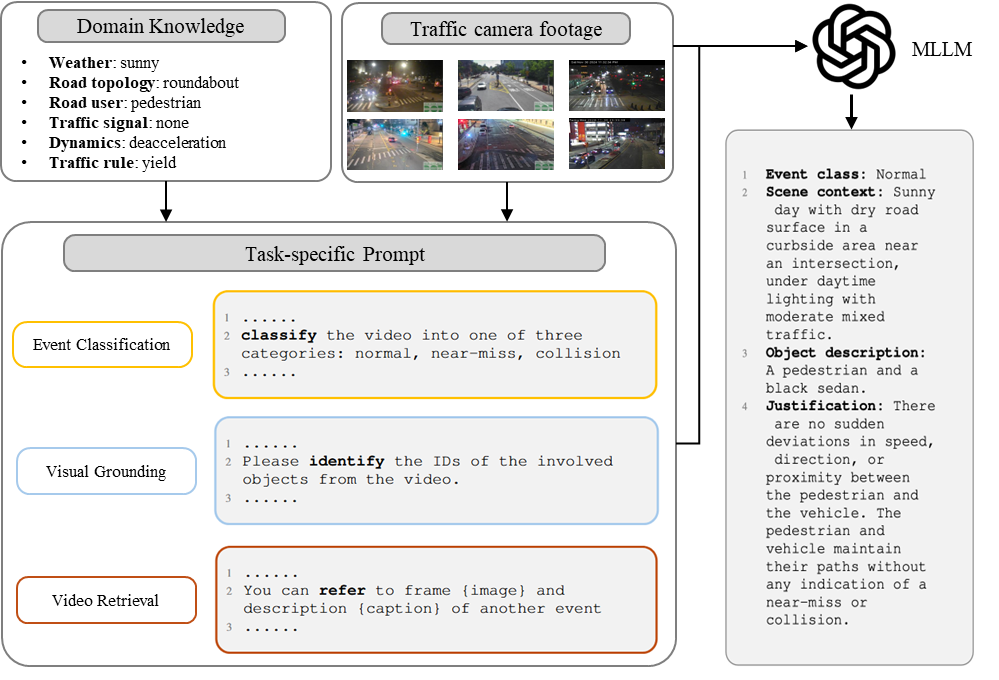}
	\caption{We aim to leverage Multimodal Large Language Models (MLLMs) to assist traffic accident analysis. Domain knowledge and traffic camera footage construct the task-specific prompt. The MLLMs generate structured responses.}
	\label{fig: introduction/overview}
\end{figure}

The emergence of Multimodal Large Language Models (MLLMs) has opened up exciting directions due to their ability to process complex data across multiple modalities (vision, sound, and more) and produce coherent, text-based outputs in a human-like logic. Before MLLMs, the potential of Large Language Models for transportation has been explored in motion planning \cite{yang2023llm4drive}, traffic rule understanding \cite{zheng2023trafficsafetygpt}, and data management \cite{zhang2024trafficgpt}. However, these Large Language Models can only take text inputs and thus are incapable of tasks requiring visual inputs. Visual input is crucial for transportation tasks as it provides rich, real-time contextual information that is much more available than textual data. For example, there is no textual equivalent of a traffic camera feed or a driver’s facial expression during a critical moment. Recently, researchers have used MLLMs with visual input processing to apply image captioning and video question-answering to transportation problems. Zhang et al. \cite{zhang2024integrating} employ MLLMs to classify driving behaviors and provide risk assessment using images from in-cabin cameras. MLLMs also show remarkable performance in scene understanding and causal reasoning for autonomous driving \citep{wen2023road, cui2024survey} and traffic event analysis \citep{wang2024accidentgpt, abu2024using, zhou2024gpt}. Although early efforts show the potential of multi-layered learning models (MLLMs) in understanding traffic scenes, they primarily concentrate on vehicle-centric reasoning tasks. These tasks are designed to enhance "surrounding awareness" for subsequent activities, such as motion planning, which are not directly applicable to traffic monitoring. In contrast, we leverage MLLMs to primarily assist traffic managers and researchers in efficiently processing large-scale video data with enhanced accessibility and reduced effort. Our motivation is straightforward: the more traffic accidents we find, the more we can analyze and the better we can prevent them in the future. 

We address this problem by going beyond simply using MLLMs as a standalone module. Instead, we propose a generalized framework that augments MLLMs with additional functionality focusing on traffic accident analysis within the scope of accident-aware video classification and visual grounding, as shown in Fig. \ref{fig: introduction/overview}. Imagine when a traffic safety engineer wants to retrieve accident videos from thousands of traffic cameras streaming 24/7 to analyze their spatial distribution. Using our methodology, this analyst could simply send a request in plain language: \textit{"Find those containing accidents at intersections and highlight the involved objects from the provided videos..."} and a program would parse this request, pass through the candidate videos, explain the accident's existence, and finally return the qualified videos, if any. However, this inspiring example has yet to become a reality due to three major concerns: (i) \textbf{scalability}: the video understanding performance of current MLLMs degrades as video duration increases. Traffic incidents, such as collisions and near-misses, are rare events embedded within lengthy footage, making it impractical to process long videos and identify accidents all at once; (ii) \textbf{unstructured responses}: the existing applications of MLLMs usually generate textual responses in a single chunk where attributes (scene environment and object dynamics) of the queried video are mixed together. The absence of structured outputs creates a post-processing burden for attribute-level retrieval; and (iii) \textbf{fine-grained analysis}: by directly inputting raw video data, MLLMs struggle to capture the fine-grained details needed for tasks such as visual grounding (identifying involved road users in the risky events), restricting their ability to going beyond scene-level understanding. Our work addresses these challenges with an emphasis on improving scalability, manageability, and extensibility. Instead of requiring MLLMs to process full-length videos, we introduce a flexible framework that allows users to split videos into clips and apply a severity-based aggregation mechanism to fuse knowledge from each clip. Task-specific multimodal prompts are designed to instruct MLLMs to produce structured, easily queryable outputs via textual prompts, while visual prompts augment video inputs with task-specified annotations, enabling fine-grained tasks such as object-level visual grounding. We have also observed that the existing natural language processing metrics are incapable of evaluating alignment with structured safety-critical textual responses, highlighting the need for a metric tailored for accident analysis. We summarize the key contributions of this work as follows:

\begin{itemize}
    \item We propose a novel MLLM-integrated framework, SeeUnsafe, to assist in traffic accident analysis, marking us one of the first attempts in this domain.
    \item The severity-based aggregation scheme is designed to handle videos of various lengths with user-defined temporary granularity.    
    \item We design multimodal prompts to guide MLLMs in generating structured content for easier querying, evaluation, and fine-grained visual grounding in accident analysis.
    \item An MLLM-based metric is proposed to score textual response quality by matching key information, and it shows better robustness and alignment in safety-critical evaluations.
    \item Experiments on the Toyota Woven Safety dataset demonstrate the effectiveness of our framework in accident analysis tasks, including accident-aware classification, object grounding, and infrastructure-vehicle co-reasoning. The code will be publicly available.
\end{itemize}
\section{Related works}
\subsection{Multimodal Large Language Models in Intelligent Transportation Systems}
Large language models (LLMs), based on the Transformer architecture \cite{vaswani2017attention}, have demonstrated advanced reasoning capabilities and have been applied to autonomous driving \cite{yang2023survey}, traffic safety question answering \cite{zheng2023trafficsafetygpt}, and traffic data management \cite{zhang2024trafficgpt}. Multimodal Large Language Models (MMLMs) extend these capabilities to image and video data, focusing on tasks like captioning and question-answering. For example, DDLM \cite{zhang2024integrating} uses a fine-tuned MLLM, LLaVA \cite{liu2023llava}, to detect driver behaviors from in-vehicle images, while \cite{de2023llm} combine MLLMs for reasoning over time series and image data to enhance autonomous driving systems. Recent works \cite{wang2023accidentgpt, zhou2024gpt} investigate using ChatGPT-4V traffic accidents using sequential image inputs. However, existing studies largely focus on vehicle-centric surrounding awareness, and few explorations have been made for traffic accident analysis from a city management perspective. To address this gap, we propose SeeUnsafe, which leverages MLLM agents to enable user-friendly interactions for accident-aware video classification and identifying objects involved in critical events, empowering traffic managers with cutting-edge AI tools.

\subsection{Vision-based traffic event analysis}
The rapid advancement of computer vision technologies renovates the data collection strategies in the era of Intelligent Transportation Systems (ITS). Compared to traditional data sources of loop detectors \cite{xu2013predicting, dia2011development}, cell phones \cite{shlayan2016exploring}, social media \cite{gu2016twitter}, and official reports \cite{janstrup2016understanding}, images and videos contain contextual information and allow for detailed incident analysis \cite{sivaraman2013looking}. Object detection and tracking algorithms are capable of identifying target objects and tracking their movement across successive frames. \cite{ijjina2019computer, fu2019investigating} utilize these algorithms to track subject vehicles within a scene, enabling trajectory-based conflict analysis. Semantic segmentation assigns one label to pixels, enabling the understanding of images at a pixel level and differentiation between various objects and regions within the image. Such techniques have empowered scene-oriented information processing, including road surface conditions in special weathers \cite{liang2019winter, abdelraouf2022using}, detecting road cracks and defects \cite{nagaraj2022semantic, shim2020lightweight}, and identifying road topology \cite{mattyus2017deeproadmapper}. Image classification is also used to estimate road weather \cite{dahmane2018weather, xiao2021classification} to support the scene-oriented incident analysis. Beyond using standard computer vision technologies for traffic footage, the rise of multi-modal visual processing opens new possibilities for open-vocabulary traffic incident analysis. Video text retrieval aims to find the best-matched videos using text queries and vice versa. \cite{nguyen2022text} proposes a vision encoder and text decoder architecture for retrieving video footage of target vehicles using text queries. Built upon the realization of video text retrieval, video question answering (VQA) \cite{ma2024robust} and video captioning \cite{liu2023traffic, dinh2024trafficvlm}, which analyze and describe video content, respectively, are another two newly introduced multi-modal tasks to the transportation community. Traditional vision-based traffic incident analysis relies on task-specific computer vision algorithms and requires post-processing to derive safety measures. In contrast, multi-modal vision models directly translate raw visual inputs into natural language outputs. However, their inference process operates in a regression manner and lacks interoperability, posing risks to safety-critical analysis.

\subsection{Datasets for traffic safety research}

Critical traffic events, such as collisions and near-misses, are rare, making real-world datasets scarce and difficult to collect. This is exactly the same problem that motivated us to do this work. Recent efforts on curating datasets have been made for both simulation and real-world datasets. CLEVRER \cite{yi2019clevrer} focuses on collision event reasoning among several simple visual objects using fully synthetic videos. DeepAccident \cite{wang2024deepaccident} creates a predefined vehicle-centric accident prediction benchmark for autonomous driving. Rank2tell \cite{sachdeva2024rank2tell} is a multi-modal ego-centric dataset for improving driving scenario understanding dedicated to advanced driver assistance systems (ADAS). The recently released Toyota Woven Traffic Safety Dataset (WTS) \cite{kong2024wtspedestriancentrictrafficvideo} focuses on pedestrian-centric traffic video understanding, with each event recorded from multiple views. Compared to other datasets, the WTS dataset has fine-grained visual details regarding the pedestrian-vehicle interaction and, therefore, is more suitable for vision-based traffic accident analysis.

\section{Methodology}
\label{sec: methodology}

\begin{figure}[t]
	\centering
	\includegraphics[width=.9\textwidth]{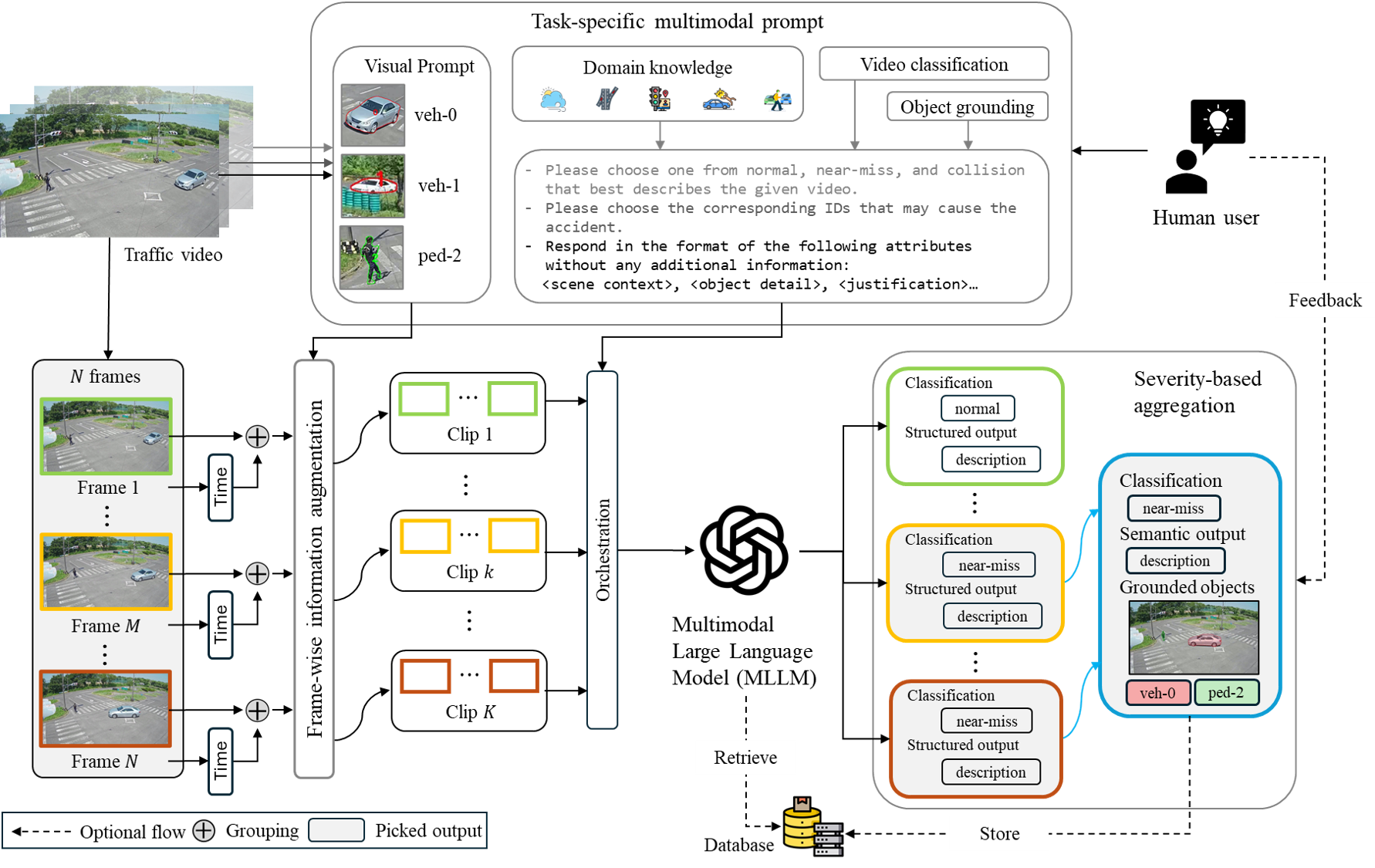}
	\caption{The overall architecture of our SeeUnsafe framework.  It processes video inputs by generating visual prompts for N frames and splitting them into K clips. Textual prompts guide the MLLM agent in generating structured responses for each clip. Outputs are then aggregated using a severity-based protocol to deliver overall event classification, semantic descriptions, and object grounding in critical events.}
	\label{fig: methodology/framework}
\end{figure}

In this section, we introduce SeeUnsafe, an MLLM-integrated framework, to assist accident-aware video classification, as shown in Fig. \ref{fig: methodology/framework}. We first give the formal definition of our task in Section \ref{sec: methodology/problem_statement}. Then, we illustrate two plug-and-play modules: severity-based aggregation in Section \ref{sec: methodology/aggregation}  and task-specific multimodal prompt in Section \ref{sec: methodology/prompt}, respectively. Lastly, we show the limitations of the existing natural language metrics for structured responses in accident scene understanding and propose the MLLM-based Information Matching Score (IMS) in Section \ref{sec: methodology/ims}.

\subsection{Problem statement}
\label{sec: methodology/problem_statement}
We formally define the accident-aware video classification problem discussed in this work. Denote an input traffic video of $N$ frames as $V = \{I_i\}^N_{i=1}$ and assume that the video contains exactly one type of event: a collision, a near-miss, or normal activity. We refer to collisions and near-miss as \textit{critical events} and are constrained to occur exclusively between two road users. The goal is to classify each video into one of three categories: normal activity, near-miss, or collision, following the standard definition of video classification \citep{karpathy2014large}. For videos classified as near-miss or collision, the secondary task is to identify the two road users involved, known as visual grounding \citep{peng2023openscene}. The target outputs include a single video class label and the identities of the two involved road users. Labels ${0,1,2}$ represent normal activity, near-miss, and collision, respectively. Multimodal Large Language Model (MLLM) agents are denoted as $A$, with prompts represented by $P$. When querying a video, the MLLM agents produce outputs such as the predicted class label $\hat{y}$, textual responses $\hat{d}$, and object identities $\hat{\boldsymbol{o}}$. Superscripts are used to distinguish instances of the same symbol where needed. For clarity, we treat video and event as equivalents in this work and use the terms interchangeably in the following sections.

\subsection{Severity-based aggregation}
\label{sec: methodology/aggregation}
Capturing temporal dependencies in videos remains a significant challenge for MLLM agents, primarily due to the curse of dimensionality inherent in visual inputs, such as resolution and temporal range. This limitation is particularly pronounced in traffic monitoring applications, where accidents and near-misses are rare events embedded within vast amounts of daily footage. The natural idea is to split long videos into shorter clips to analyze them individually within the MLLM agents' capability. With possibly many shorter clips, some of them may be classified as critical events (collision or near-miss), while others may contain only normal activities. A simple majority voting approach \citep{hasan2024vision} may suppress rare critical events, and one should consider classifying videos by the most severe event across all clips, prioritizing collisions and near-misses. Based on this, we propose a severity-based aggregation method to fuse classification results from multiple clips, as illustrated in blue in Fig. \ref{fig: methodology/framework}. Without losing generality, we assume the existence of a splitting function $g$ that divides the total $N$ frames chronologically into $K$ non-overlapping clips $C_k$ such that:
\begin{align}
    &\{C_k\}^K_{k=1} = g(\{I_i\}^N_{i=1})\\
    &\{I_i\}^N_{i=1} = \bigcup^K_{k=1} C_k\\
    & C_m \cap C_n = \emptyset, \ \forall m\neq n\\
    &C_{1,1} = I_1\\
    &C_{K, |C_K|} = I_N
\end{align}
where $C_{k,j}$ indicate the $j$-th frame in the $k$-th clip. By sending $K$ clips parallel into the MLLM agent $A$ with prompt $P$, the set of output tuples is:
\begin{align}
    \big\{(\hat{y}_k, \hat{d}_k)\big\}^K_{k=1} = \big\{A(P, C_k)\big\}^K_{k=1}
\end{align}
We use the label value as a proxy for event severity, with collisions (2) being the most severe and normal activity (0) being the least severe. The video $V$ is classified based on the highest severity class among all clips. Descriptions from clips with the highest severity are aggregated using a fusion function $h$. The final classification label $\hat{Y}$ and responses $\hat{D}$ for video $V$ are then obtained as:
\begin{align}
    \hat{Y} &= \arg\max\big(\{\hat{y}_k\}^K_{k=1}\big)\\
    \hat{D} &= h\Big(\big\{\hat{d}_k\cdot \mathbbm{1}\big(\hat{y}_k=\hat{Y})\big\}^K_{k=1}\Big)
\end{align}

This aggregation method integrates seamlessly with any user-defined splitting function $g$ and fusion function $h$. For example, when $K=1$, the entire video is processed by the MLLM agent $A$ in a single pass, assuming token capacity permits. When $K=N$, the method aggregates $N$ image-wise reasoning responses, degenerating to many existing image-based traffic scene understanding works. In our experiments, we use a uniform splitting function $g$ and a heuristic fusion function $h$. The choice of $g$ and $h$ is a design art and is beyond the scope of this work. We provide further discussion in Section \ref{sec: discussion}. Visual grounding for object identification involved in critical events occurs after the severity-based aggregation and will be detailed in the following Section \ref{sec: methodology/prompt}.
 
\subsection{Task-specifc multimodal prompt}
\label{sec: methodology/prompt}

\begin{figure}[t]
	\centering
	\includegraphics[width=.8\textwidth]{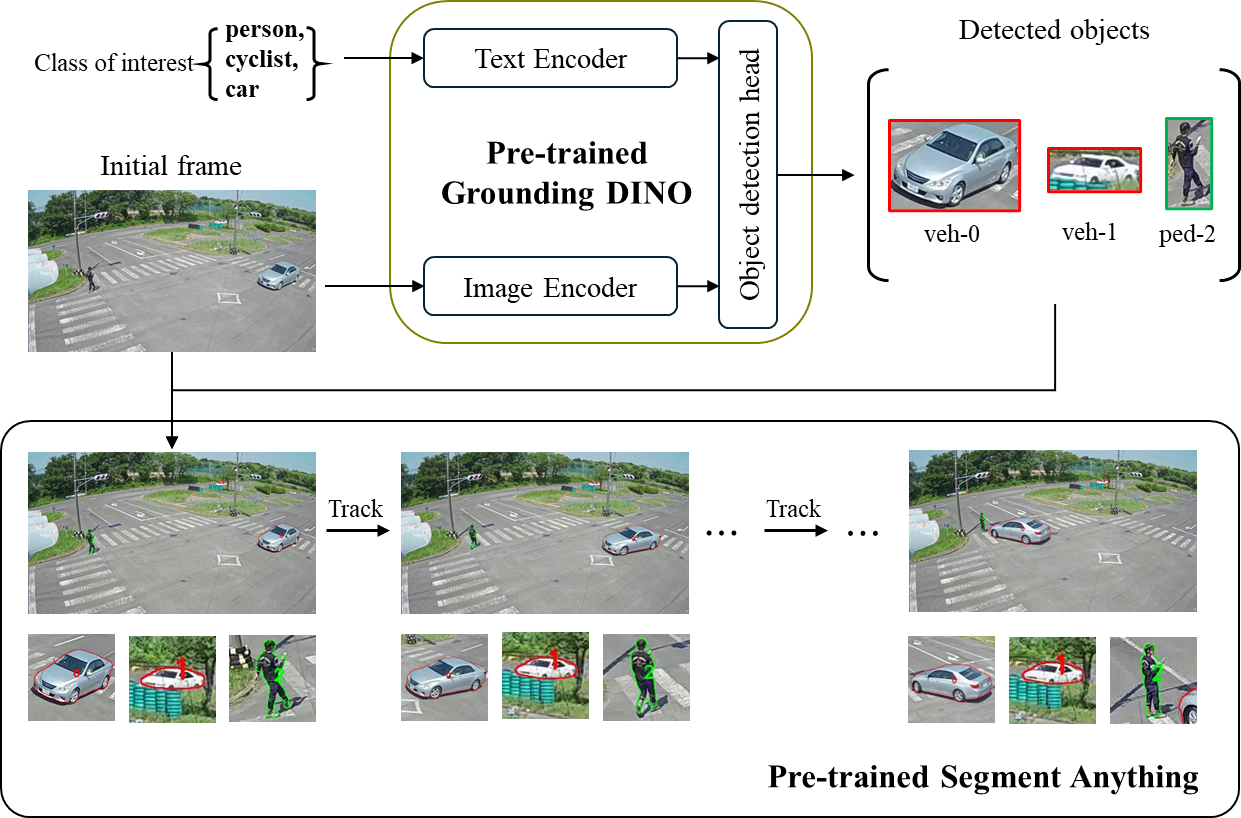}
	\caption{Workflow for generating visual prompts. An open-set object detection model identifies the bounding boxes of objects of interest and assigns a unique ID to each object in the given frame. These bounding boxes are then used to guide segmentation models, which generate instance masks and track objects across frames. Only the object contours are overlaid, while the complete masks are stored for future indexing.}
	\label{fig: methodology/visual_prompt}
\end{figure}

Prompts are crucial in MLLM applications as they provide a prior that conditions the MLLM to generate posterior responses. A more informative prior leads to better posterior outputs. In this work, we aim not only to predict the video class but also to generate responses that justify the prediction—an advantage MLLMs have over traditional video classification models. Like multimodal inputs enrich the information sources, prompts can also be in the format of multiple modalities that bring various priors \citep{khattak2023maple, jiang2022vima}. The prompt used in this work consists of both textual and visual components. For the textual component, the design is guided by the question: "What attributes can describe a traffic event?" Traffic event descriptions can include several attributes: scene context (e.g., weather, road topology, lighting conditions), object information (e.g., types of objects in the video and their appearance), and justification (e.g., the explanation for classifying the event into a specific category). These attributes define the desired output responses from MLLM agents. Specifically, we ask the MLLM agent to generate structured responses across four predefined attributes: \texttt{event class}, \texttt{scene context}, \texttt{object description}, and \texttt{justification}. Structuring responses in this way leads to more precise descriptions and can slightly improve classification accuracy, as we will show in Section \ref{sec: experiment}. Moreover, these structured responses facilitate attribute-based retrieval, as discussed in the use case in Section \ref{sec: introduction} and illustrated with a dashed arrow in Fig. \ref{fig: methodology/framework}.

In addition to textual prompts, visual prompts have recently emerged as a powerful tool for assisting MLLMs in fine-grained vision tasks \citep{wang2023review, jia2022visual}. Visual prompts provide task-specific annotations or hints, such as highlighting important regions of an image or video, helping the model focus on areas of interest or resolve task ambiguities. In this work, we use object-wise segmentation masks as visual prompts to identify objects potentially involved in critical events. During pre-processing, visual prompts are added to the original images before being sent to the MLLM agent. The procedure is illustrated in Fig. \ref{fig: methodology/visual_prompt}. Specifically, we take the first frame of the video and apply GroundingDINO \citep{liu2023grounding} for open-vocabulary object detection, allowing the detection of arbitrary objects based on language descriptions. The classes of interest are predefined persons, cyclists, and cars, which are the most common road users in daily traffic. Detected objects (bounding boxes and track IDs) are then passed to Segment Anything \citep{kirillov2023segany}, which tracks each object across the video and generates segmentation masks, as shown in Fig. \ref{fig: methodology/visual_prompt}.

To identify road users involved in the critical events, we use images augmented with the visual prompt from clips that belong to the same class as the predicted video class $\hat{Y}$, denoted as ${\hat{I}}$. The corresponding textual prompt for visual grounding is denoted as $P^{grounding}$. Visual prompts are indexable for each detected object, including their masks and track IDs. Given the predicted video class $\hat{Y}$, the MLLM visual grounding agent $A^{grounding}$ identifies the track IDs $\hat{\boldsymbol{o}}$ as follows:
\begin{align}
    &\hat{\boldsymbol{o}} = A^{grounding}(P^{grounding}, \{\hat{I}\})\\
    &\{\hat{I}\} = \bigcup^K_{k=1} C_k\cdot \mathbbm{1}\big(\hat{y}_k=\hat{Y})
\end{align}

In this work, we use object boundaries as visual prompts for two key reasons. First, masked areas are smaller than bounding boxes, allowing boundaries to implicitly capture object poses through their contours. Second, segmentation masks derived from boundaries can be indexed for visual grounding, providing more intuitive visualization compared to bounding boxes. While overlaying object boundaries enables finer-grained visual grounding, it can also introduce noise into the image. To assess the impact of these visual prompts on video classification performance and textual responses, we conduct extensive experiments, as detailed in Section \ref{sec: experiment}.

\subsection{Information matching score}
\label{sec: methodology/ims}

\begin{figure}[ht]
	\centering
	\includegraphics[width=.8\textwidth]{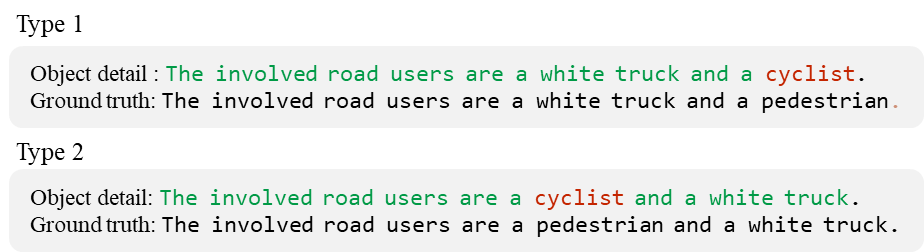}
	\caption{We show two types of potential failures in structured responses for the \texttt{object\_description} component when describing road users involved in a collision. Correct contents are highlighted in \textcolor{black!40!green}{green}, while incorrect ones are in \textcolor{black!20!red}{red}. (Best viewed in color and see also Table \protect\ref{table: methodology/nlp_metric}).}
	\label{fig: methodology/nlp_metric}
\end{figure}

In addition to evaluating classification performance, it is crucial to assess the correctness of the output responses, as these responses may be used for decision-making and retrieval indexing. In the natural language processing (NLP) domain, metrics such as Bilingual Evaluation Understudy (BLEU) and Recall-Oriented Understudy for Gisting Evaluation (ROUGE) are widely used to evaluate text quality. However, these metrics primarily focus on the similarity between predicted sentences and ground truth, often overlooking nuanced yet critical mismatches. This limitation makes them unsuitable for traffic accident analysis, where even a single word can drastically alter the justification or interpretation of an event. To illustrate this issue, we provide a motivating example in Fig. \ref{fig: methodology/nlp_metric} that highlights the shortcomings of naively applying conventional NLP metrics to traffic accident analysis.


\begin{table}
\centering
\caption{Evaluation results of the motivating example using BLEU and ROUGE.}\label{table: methodology/nlp_metric}
\begin{tabular}{lllll}
\toprule
Metric & BLEU & ROUGE-1 & ROUGE-2 & ROUGE-L\\
\midrule
Type 1 & 0.89 & 0.91 & 0.90 & 0.91 \\
Type 2 & 0.70 & 0.91 & 0.90 & 0.91 \\
\bottomrule
\end{tabular}
\end{table}

Two key observations emerge from this example: (1) Although \texttt{object\_detail} and \texttt{ground\_truth} convey the same semantic information, merely switching the position of the error term within the sentences reduces the BLEU score from 0.89 to 0.70, a 21.3\% drop; (2) Despite incorrectly identifying the involved object as a cyclist instead of a pedestrian, the ROUGE score remains no less than 0.90, suggesting satisfactory matching performance. In summary, despite the critical error of confusing a pedestrian with a cyclist, the BLEU and ROUGE scores remain relatively high, highlighting their inability to characterize failure contexts. Thus, these metrics are unreliable for accident analysis and evaluation.

Given the limitations of traditional NLP metrics in accident analysis, we propose a new metric, the Information Matching Score (IMS). Humans can easily identify critical errors while ignoring minor mismatches, understanding that some information (e.g., road topology, lighting conditions, object classes, and dynamics) is more important than others. However, due to the inherent flexibility of language, evaluating information matching across varied narratives is challenging. IMS leverages the information understanding and extraction capabilities of MLLMs to mimic human behavior when comparing texts in the context of traffic accident analysis. As shown in Prompt~\ref{lst:visual_perception_prompt}, we design an attribute-matching prompt to instruct the MLLM agent to compare the information-matching level between generated responses and ground truth for each attribute of \texttt{scene\_conetext}, \texttt{object\_description}, and \texttt{justification}. The MLLM then generates component-wise matching scores in a 100-scale to evaluate VLM output quality. To avoid the "modal collapse" phenomenon and enhance evaluation robustness, we perform $T$ trials with a non-zero temperature $\eta$, which controls output randomness and creativity \citep{peeperkorn2024temperature}: higher temperature values result in more random and creative outputs.

\begin{lstlisting}[language=Python, caption=Information Matching Score, label=lst:visual_perception_prompt]
## Prompt MLLM to evaluate the matching quality
- You are tasked with evaluating the overall matching quality between the ground truth and predicted descriptions for three components: scene context, object description, and justification.
- Please carefully consider the latent correlations among these components and provide individual scores for each.
- 1. (*@\bf{Scene\_Context:}@*) Focus on the factors such as weather, lighting conditions, and road topology (i.e. intersection). These are representative factors to categorize the analyzed video. For missing content in the predicted descriptions, consider them as wrong.
- 2. (*@\bf{Object\_Description:}@*) Focus on the factors including number of objects, the class of objects, and appearance information. For missing content in the predicted descriptions, consider them as wrong.
- 3. (*@\bf{Justification:}@*)  This is the most important component. Be strict and rigorous when evaluating this component. Pay attention to descriptions regarding the dynamics, movement, and relative positions of the involved objects. If you think the predicted description misses important contents regarding the ground truth or incorrect, assign a low score.
- Please provide a score from 0 to 100 for each component.
- 100: Perfect alignment with no missing or incorrect information.
- 80(*@-@*)99: Very good alignment with only minor discrepancies.
- 60(*@-@*)79: Good alignment but with some significant missing or incorrect details.
- 40(*@-@*)59: Moderate alignment with noticeable errors or omissions.
- 20(*@-@*)39: Poor alignment with many incorrect or missing details.
- 0(*@-@*)19: Completely incorrect or irrelevant prediction.
\end{lstlisting}

Let the IMS for the $t$-th trial of the $l$-th attribute be $s_t^l$, where $t \in {1, 2, \cdots, T}$ represents $t$-th trials and $l \in {1, 2, \cdots, L}$ represents $l$-th response attribute in the generated response. IMS is obtained following Eq. (\ref{eq: infomatch}, \ref{eq: score}), where $\alpha_l$ is the weight of the $l$-th attribute, $A^{IMS}$ is the MLLM agent for generating IMS, $\hat{D}^l$ is the $l$-th attribute in the generated response, $D^l$ is the $l$-th attribute in the ground truth, $\eta$ is the temperature, and $P^{IMS}$ is the textual prompt used for computing IMS. We compare IMS with BLEU and ROUGE-L and show its better alignment with human common sense in Section \ref{sec: experiment}.
\begin{align}
   \label{eq: infomatch}
    &\text{IMS} = \frac{ \alpha_l}{T} \sum^L_{l=1} \sum^T_{t=1} s^l_t\\
    & s^l_t = A^{IMS}(\hat{D}^l, D^l, \eta, P^{IMS})
    \label{eq: score}
\end{align}

\section{Experiments}
\label{sec: experiment}
\subsection{Implementation details}
\begin{figure}[htbp]
	\centering
	\includegraphics[width=0.7\textwidth]{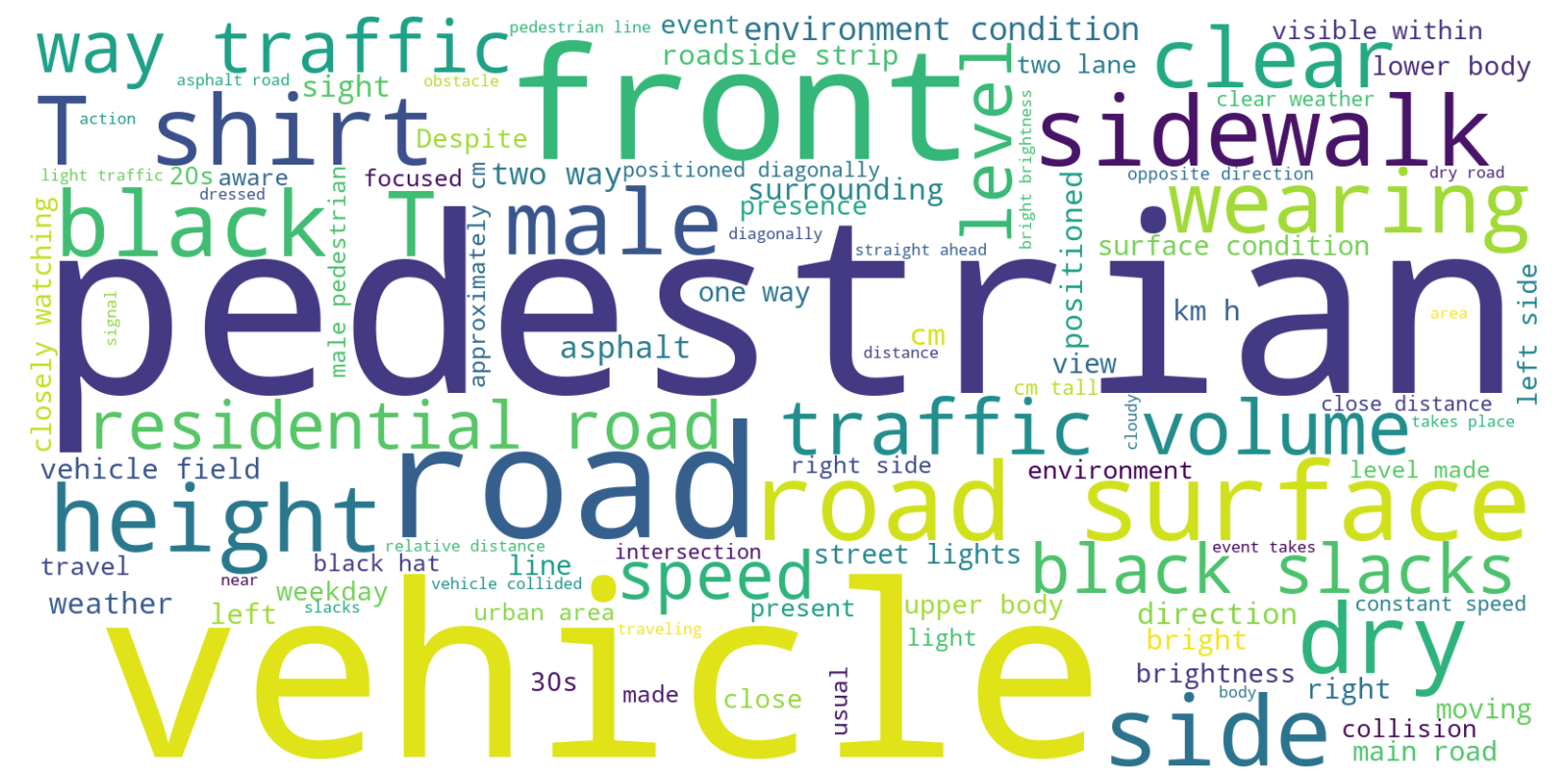}
	\caption{Word cloud of the 100 most frequent words in the processed WTS dataset captions. Larger words appear more frequently.}
	\label{fig: experiment/word cloud}
\end{figure}

We conduct experiments on the Toyota Woven Traffic Safety (WTS) dataset \citep{kong2025wts}, a real-world collection focused on pedestrian-vehicle interactions, originally designed for video captioning tasks. The dataset includes 249 scenarios (167 in training and 82 in validation). In this work, results are reported on all 249 scenarios to explore the feasibility of accident-aware video classification and visual grounding. Each scenario provides one to four videos from overhead surveillance cameras and one video in-vehicle dash camera. During the data preprocessing, we manually select the best PoV from the overhead videos in one scenario. The ground truth consists of captions describing pedestrian and vehicle behaviors, which we summarize into four predefined attributes, event class, scene context, object description, and justification using GPT-4o, followed by manual verification for accuracy. Detailed dataset statistics are provided in Appendix \ref{appendix: dataset}, and a word cloud of the ground truth captions we use is shown in Fig. \ref{fig: experiment/word cloud}.

Following the mainstream setting for video understanding tasks \citep{goyal2017something, lin2019tsm}, we use 9 frames uniformly sampled from the videos and split them into 3 clips of 3 frames each. If multiple clips belong to the highest severity class, we select the response from the earliest clip. We compare three MLLM agents, LLaVA-NeXT \citep{li2024llava}, GPT-4o mini \citep{openai2024gpt4omini}, and GPT-4o \citep{openai2024gpt4o}, and one non-MLLM model, VideoCLIP \citep{xu2021videoclip}. For reproducibility, the temperature is set to 0 for GPT-4o and GPT-4o mini, and 0.1 for LLaVA. We use Segment Anything and GroundingDINO with default hyperparameters, and modify the visual prompt generation process from \citep{wang2024vlm}. IMS is computed using a GPT-4o agent with a temperature of 0.5, repeated three times for average. Visual grounding is successful only if both road users are correctly indexed and any errors are considered a failure. "VP" stands for "visual prompt." in this section. The complete textual prompts used in this work are provided in Appendix \ref{appendix: prompt}.

\subsection{Quantitative analysis}
We compare SeeUnsafe, powered by a GPT-4o agent, with state-of-the-art MLLMs, including LLaVA, GPT-4o, and a lighter version, GPT-4o mini. Models labeled with "vanilla" (e.g., GPT-4o (vanilla), GPT-4o mini (vanilla)) process all nine frames simultaneously without added visual prompts or splitting and use a simplified version of the SeeUnsafe prompt (excluding instructions for structured output). As shown in Table \ref{tab: experiemnt/main_result}, the traditional VideoCLIP model achieves the lowest scores for video classification and cannot generate responses due to its design, highlighting the limitations of such vision-language pre-trained models for this task. During the experiment, we observed that while $\text{LLaVA-NeXT}_{video}$ could generate some responses, it consistently predicted the same video class regardless of the actual class (see the last column in Fig. \ref{fig: experiment/confusion_matrix_main}). As a result, we excluded it from the accident-aware classification evaluation. Among the remaining models, GPT-4o (vanilla) performed second-best across most metrics, with its lighter version, GPT-4o mini (vanilla), performing slightly worse but still outperforming the other models. This trend highlights that overall performance improves with the reasoning capabilities of MLLMs. Notably, SeeUnsafe is the only one capable of performing visual grounding credit to the added visual prompt, while all other models fail at this task.

\begin{table}
\centering
\caption{Performance of accident-aware video classification and visual grounding. The best and the second best results are shown in \textbf{bold} and with \underline{underline}, respectively. $\text{SR}_{gronding}$ is the visual grounding success rate.}
\label{tab: experiemnt/main_result}
\begin{threeparttable}
\begin{tabular}{llllllll}
\toprule
Model       & Accuracy & Precision & F1-Score  & $\text{SR}_{gronding}$ & BLEU\tnote{*} & ROUGE-L\tnote{*} & IMS\tnote{*} \\ \midrule
VideoCLIP   &  27.71  &  20.98  &  22.38  &  -  &  -  &  -  &  -  \\
$\text{LLaVA-NeXT}_{video}$  &  -  &  -  &  -  &  -  &  14.14  &  11.50  &  15.12  \\
GPT-4o mini (vanilla) &  58.23  &  69.38  &  62.04  &  -  &  21.36  &  14.98  &  21.27  \\
GPT-4o (vanilla)     &  \underline{71.49}  &  \underline{70.62}  &  \underline{70.92}  &  -  &  \underline{24.69}  &  \underline{16.90}  &  \underline{23.47}  \\
SeeUnsafe (Ours)  &  \textbf{76.31}  &  \textbf{72.00}  &  \textbf{72.62}   &  \textbf{51.47}  &  \textbf{37.42}  &  \textbf{37.72}  & \textbf{43.41}   \\
\bottomrule
\end{tabular}%
\begin{tablenotes}
  \item[*] Equally-weighted average over three components.
\end{tablenotes}
\end{threeparttable}
\end{table}

\begin{figure}[t]
	\centering
	\includegraphics[width=0.7\textwidth]{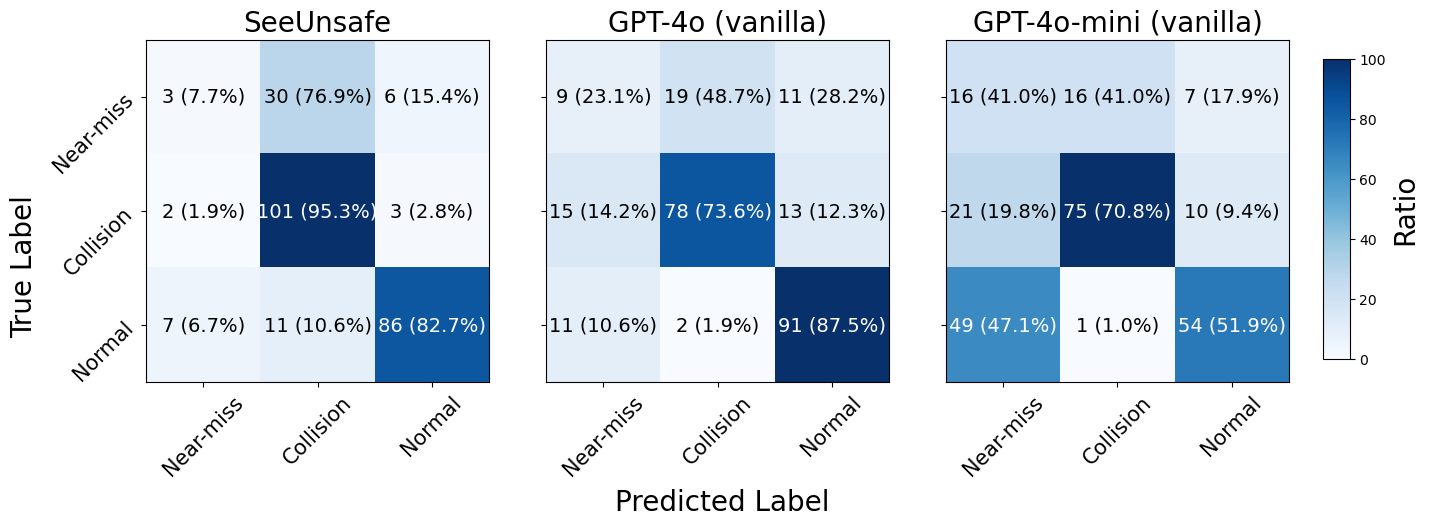}
	\caption{Confusion matrix of the SeeUnsafe and vanilla implementations of GPT-4o and GPT-4o mini.}
	\label{fig: experiment/confusion_matrix}
\end{figure}

Given that SeeUnsafe, GPT-4o (vanilla), and GPT-4o mini (vanilla) are better in accident-aware video classification, we generate confusion matrices to analyze their failure cases as shown in Fig. \ref{fig: experiment/confusion_matrix}. Since SeeUnsafe is integrated with a GPT-4o agent, their confusion matrices are similar, with SeeUnsafe performing slightly better in classifying collisions. This improvement may be due to the added visual prompt, which enhances the model's sensitivity to object boundaries and spatial interactions, leading to more videos of near-miss and normal classes being classified as collisions as well. In contrast, GPT-4o mini (vanilla), as a lighter model with weaker object interaction understanding, misclassifies most videos as collisions or near-misses, likely interpreting overlapping object areas as abnormal event indicators. We will shortly discuss the impact of different MLLM agents within the SeeUnsafe framework and the effect of visual prompts on performance.

\subsubsection{Alignment evaluation of IMS metric}

\begin{figure}[htbp]
	\centering
	\includegraphics[width=0.9\textwidth]{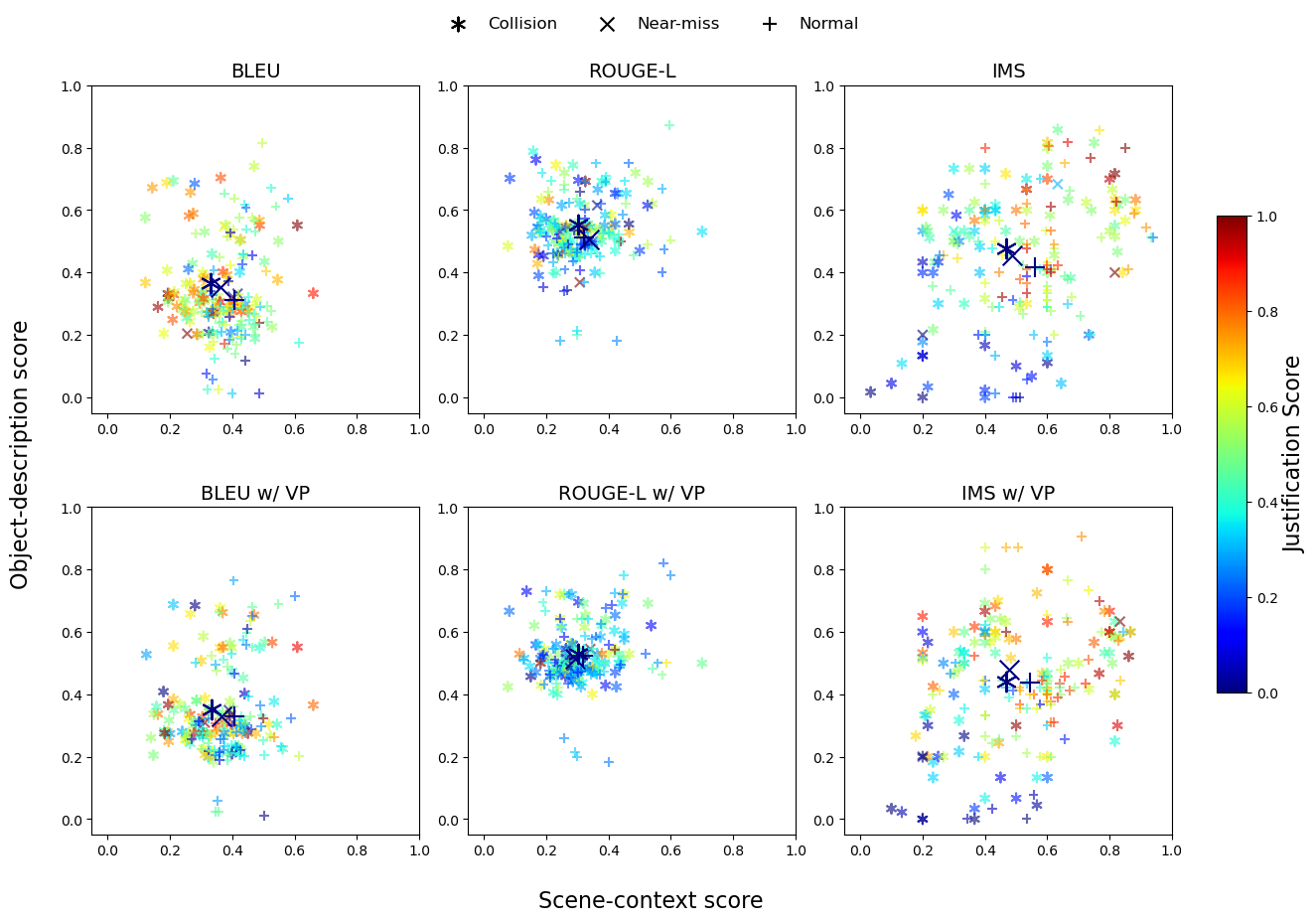}
	\caption{Visualization of semantic evaluation results of correctly classified videos with and without visual prompt based on GPT-4o. The x-axis is the \texttt{scene\_context} score, the y-axis is the \texttt{object\_description} score, and the \texttt{justification} score defines the color. There are three marker types: $*$ for collision, $\times$ for near-miss, and $+$ for normal videos, respectively. The bigger markers in black are the centroid of the corresponding video classes. We provide detailed quantitative results in Table \ref{tab: experiment/quantatitive_ims}.}
	\label{fig: experiment/llm_metric}
\end{figure}

To verify the alignment capabilities of IMS, we visualize the response evaluation results of SeeUnsafe for three attributes: scene context, object description, and justification, as shown in Fig. \ref{fig: experiment/llm_metric}. The scatter plots for BLEU and ROUGE-L show spatial clustering with uniformly distributed colors, whereas IMS values form a linear pattern from the bottom left to the top right, with brighter colors as points move upward. This indicates that the three attributes in the responses are positively correlated and collectively contribute to the final classification: better alignment in \texttt{scene\_context} and \texttt{object\_description} often leads to better alignment in \texttt{justification}, and vice versa. IMS captures this intuition more effectively, while BLEU and ROUGE-L provide less informative and sometimes misleading evaluations, as shown by the motivation example in Table \ref{table: methodology/nlp_metric}. We observe similar patterns when comparing IMS results for SeeUnsafe using MLLM agents of GPT-4o mini, $\text{LLaVA-NeXT}_{interleave}$ and $\text{LLaVA-NeXT}_{video}$, and note that stronger MLLM scorers (e.g., GPT-4o compared to GPT-4) produce stricter evaluations, while weaker scorers provide overconfident values (all additional plots can be found in Appendix \ref{appendix: evaluation ims}). Additionally, visual prompts do not significantly affect the spatial or color distributions of the scatter points. A quantitative comparison in Table \ref{tab: experiment/quantatitive_ims} highlights class-wise changes with and without adding visual prompts, corresponding to the centroids of each class in Fig. \ref{fig: experiment/llm_metric}. Across the three event classes, evaluation results remain consistent, and we discuss how visual prompts influence classification accuracy and response alignment in the next subsection.

\begin{table}[ht]
\centering
\caption{Semantic evaluations of successfully classified videos using GPT-4o. We highlight the IMS results where values in \textcolor{black!10!red}{red} indicate performance degradation relative to the same data without using visual prompts, and values in \textcolor{black!20!green}{green} indicate better performance, respectively.}
\label{tab: experiment/quantatitive_ims}
\resizebox{1\linewidth}{!}{
\begin{threeparttable}
\begin{tabular}{lllllllllll}
\toprule
\multirow{2}{*}{Class}  & \multirow{2}{*}{\begin{tabular}[c]{@{}l@{}}Visual\\ Prompt\end{tabular}} & \multicolumn{3}{c}{BLEU} & \multicolumn{3}{c}{ROUGE-L} & \multicolumn{3}{c}{IMS} \\ \cline{3-11} 
   &   & Scene & Object & Justification & Scene & Object & Justification & Scene & Object & Justification \\ \midrule
\multirow{2}{*}{Near-miss} & No & 36.23 & 35.17 & 40.56 & 33.67 & 50.73 & 31.13 & 48.75 & 45.42 & 31.25 \\
   & Yes & \textcolor{black!20!green}{36.57} & \textcolor{black!10!red}{33.02} & \textcolor{black!10!red}{36.81} & \textcolor{black!10!red}{29.35} & \textcolor{black!20!green}{51.49} & \textcolor{black!10!red}{26.00} & \textcolor{black!10!red}{47.78} & \textcolor{black!20!green}{47.78} & \textcolor{black!10!red}{30.00} \\ \cline{3-11} 
\multirow{2}{*}{Collision} & No & 33.14 & 36.57 & 37.74 & 30.11 & 55.40 & 27.32 & 46.93 & 47.69 & 26.29 \\
   & Yes & \textcolor{black!20!green}{33.31} & \textcolor{black!10!red}{35.15} & \textcolor{black!20!green}{39.49} & \textcolor{black!20!green}{30.12} & \textcolor{black!10!red}{53.01} & \textcolor{black!20!green}{27.66} & \textcolor{black!10!red}{46.69} & \textcolor{black!10!red}{44.07} & \textcolor{black!10!red}{23.56} \\ \cline{3-11}
\multirow{2}{*}{Normal} & No & 40.80 & 31.28 & 45.24 & 32.22 & 51.18 & 32.29 & 55.98 & 41.59 & 49.75 \\
   & Yes & \textcolor{black!10!red}{40.63} & \textcolor{black!20!green}{32.81} & \textcolor{black!10!red}{44.06} & \textcolor{black!10!red}{31.83} & \textcolor{black!20!green}{52.23} & \textcolor{black!10!red}{32.09} & \textcolor{black!10!red}{54.55} & \textcolor{black!20!green}{43.70} & \textcolor{black!20!green}{50.82} \\
\bottomrule
\end{tabular}%
\end{threeparttable}
}
\end{table}

\subsubsection{Performance using different MLLM agents}

\begin{figure}[htbp]
	\centering
	\includegraphics[width=0.9\textwidth]{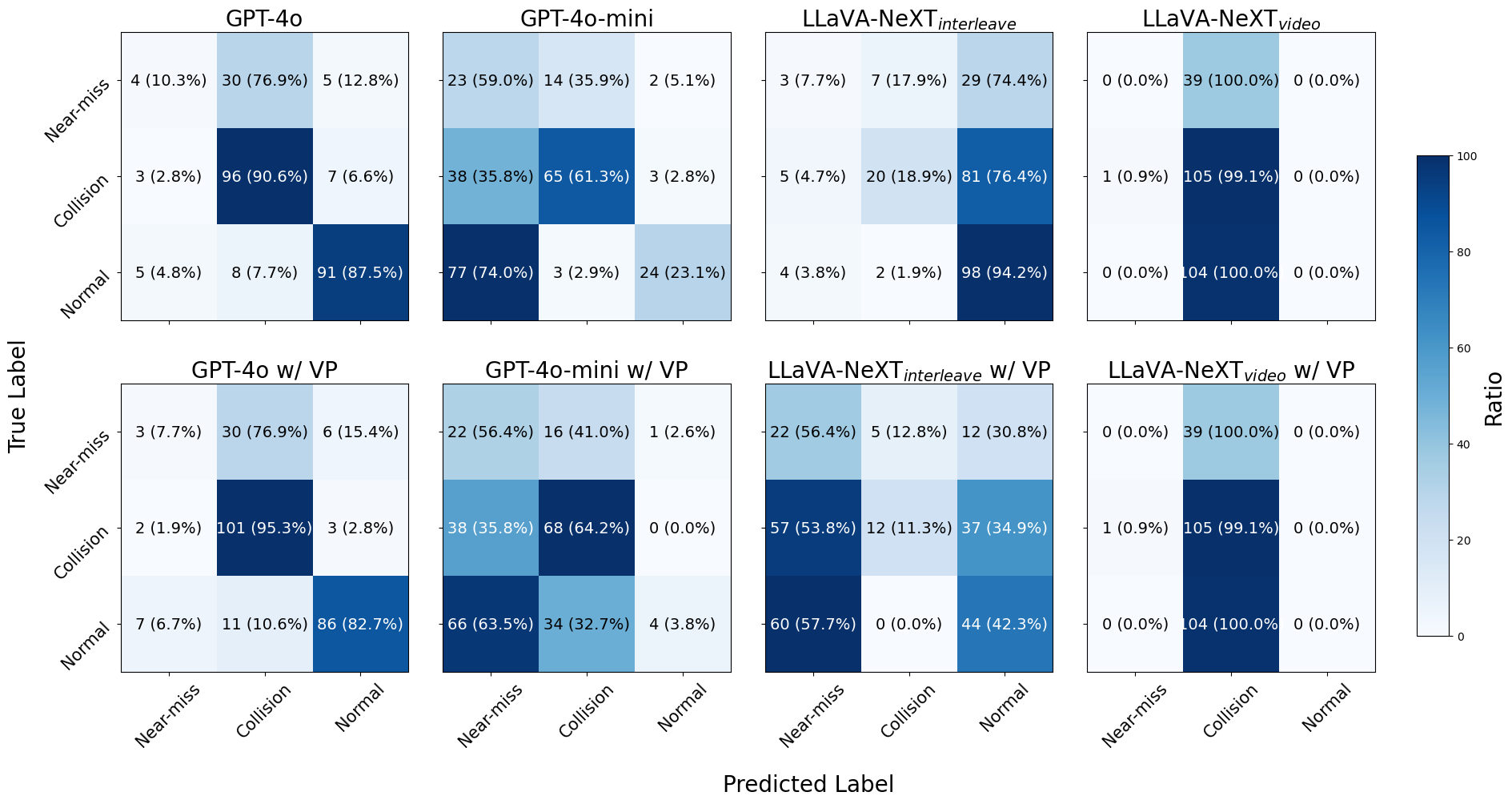}
	\caption{Confusion matrix for accident-aware video classification using different MLLM agents within the SeeUnsafe framework with and without visual prompts.}
	\label{fig: experiment/confusion_matrix_main}
\end{figure}

For our SeeUnsafe framework, MLLM agents are the core components responsible for video understanding and response generation. Thus, comparing the framework's performance with different MLLMs is essential. As shown in Table \ref{tab: experiment/mllm_comparision}, we evaluate performance with and without visual prompts. The results show that as the reasoning capability of the MLLM increases, overall performance improves, indicating that more powerful MLLM agents are preferred if resources permit. We also observe that adding visual prompts introduces a marginal performance drop with GPT-4o and a more significant drop with GPT-4o mini and $\text{LLaVA-NeXT}_{interleave}$. As mentioned in Section \ref{sec: methodology/prompt}, visual prompts introduce "noise" by adding object boundaries, which is a type of noise hardly seen by these MLLM agents during the training process. Among these MLLM agents, GPT-4o is a larger and more robust model, and it can effectively handle the added boundaries and even utilize them for visual grounding tasks. Conversely, GPT-4o mini, with fewer parameters, is more sensitive to object boundaries, resulting in classification degradation, although its response evaluations remain largely unaffected. For $\text{LLaVA-NeXT}_{interleave}$, the larger performance drop suggests greater sensitivity to visual prompts, likely due to less training data compared to the GPT-4o series. The confusion matrices in Fig. \ref{fig: experiment/confusion_matrix_main} confirm these observations. The confusion matrix of the GPT-4o agent shows minimal changes before and after adding visual prompts. While the GPT-4o mini agent experiences classification degradation, it maintains a similar pattern, excelling in collision classification but frequently misclassifying normal events as near-miss events. In contrast, $\text{LLaVA-NeXT}_{interleave}$ shows a more pronounced impact, with visual prompts causing most classifications to skew toward near-miss, regardless of the ground truth labels. Finally, the last column highlights the failure of $\text{LLaVA-NeXT}_{video}$ that uses all nine frames at a time, which produces meaningless results, classifying all collisions incorrectly.

\begin{table}
\caption{Performance of accident-aware video classification and visual grounding using different MLLM agents. The best and the second best results are shown in \textbf{bold} and with \underline{underline}, respectively. $\text{SR}_{gronding}$ is the visual grounding success rate.}
\label{tab: experiment/mllm_comparision}
\resizebox{1\linewidth}{!}{
\begin{threeparttable}
\begin{tabular}{clllllllll}
\toprule
Visual Prompt      & MLLM Module       & Accuracy & Precision & F1-Score  &  $\text{SR}_{grounding}$  & BLEU\tnote{*} & ROUGE-L\tnote{*} & IMS\tnote{*} \\ \midrule
\multirow{3}{*}{-} &  $\text{LLaVA-NeXT}_{interleave}$  &  48.59  &  52.95  &  40.69   &  -  &  20.64  &  14.51  &  25.20  \\
  &  GPT-4o mini &  44.98  &  70.92  &  48.58   &  -  &  \textbf{39.74}  &  36.92  &  36.00  \\
  &  GPT-4o & \textbf{76.71}  &  \textbf{72.62}  &  \textbf{73.24}  &  -  &  37.42  &  \textbf{38.08}  & \textbf{44.53}  \\ \hline
\multirow{3}{*}{\begin{tabular}[c]{@{}l@{}}Object  \\ Boundary\end{tabular}} & 
 $\text{LLaVA-NeXT}_{interleave}$  &  31.33  &  52.29  &  30.84   &  -  &  18.46  &  13.84  &  23.95  \\
& GPT-4o mini &  37.75  &  60.68  &  33.09   &  \underline{25.74}  &  \underline{39.64}  &  37.06  &  34.53  \\
& GPT-4o      &  \underline{76.31}  &  \underline{72.00}  &  \underline{72.62}   &  \textbf{51.47}  &  37.42  &  \underline{37.72}  &  \underline{43.41}  \\
\bottomrule
\end{tabular}%
\begin{tablenotes}
  \item[*] Equally-weighted average over three components.
\end{tablenotes}
\end{threeparttable}
}
\end{table}

\subsubsection{Performance of visual grounding}

To make a meaningful evaluation of visual grounding performance, we only consider those of collision and near-miss classes with at least three objects present to avoid trivial cases, totaling 136 videos. During manual validation, we find only 88 videos have valid masks for both involved objects and the remaining 48 have at least one involved object not correctly tracked due to failures in the detection and tracking models. However, we still report the success rate based on the total 136 videos containing critical events and note that the success rates can be easily boosted by using more robust detection and tracking models. As shown in Table \ref{tab: experiment/mllm_comparision}, using the GPT-4o agent, SeeUnsafe successfully identifies the correct road users in 70 of the 136 videos, achieving a success rate of 51.47\%, which can go up to 87.5\%  when considering only the 88 videos with valid masks. With using GPT-4o mini agent, the success rate shows a decline from 51.47\% to 25.74\% on 136 videos and from 87.5\% to 39.78\% on 88 correctly tracked videos, reflecting its reduced capacity for handling complex object interactions compared to GPT-4o. Visualization of visual grounding outputs can be found in Section \ref{sec: experiment/qualitative_Result}. These results highlight the importance of robust object tracking and GPT-4o's effectiveness in leveraging visual prompts, enabling applications like extracting trajectories for objects involved in accidents without relying on handcrafted filtering criteria.

\subsubsection{Correlation between classification accuracy and response alignment}

\begin{figure}[t]
	\centering
	\includegraphics[width=0.9\textwidth]{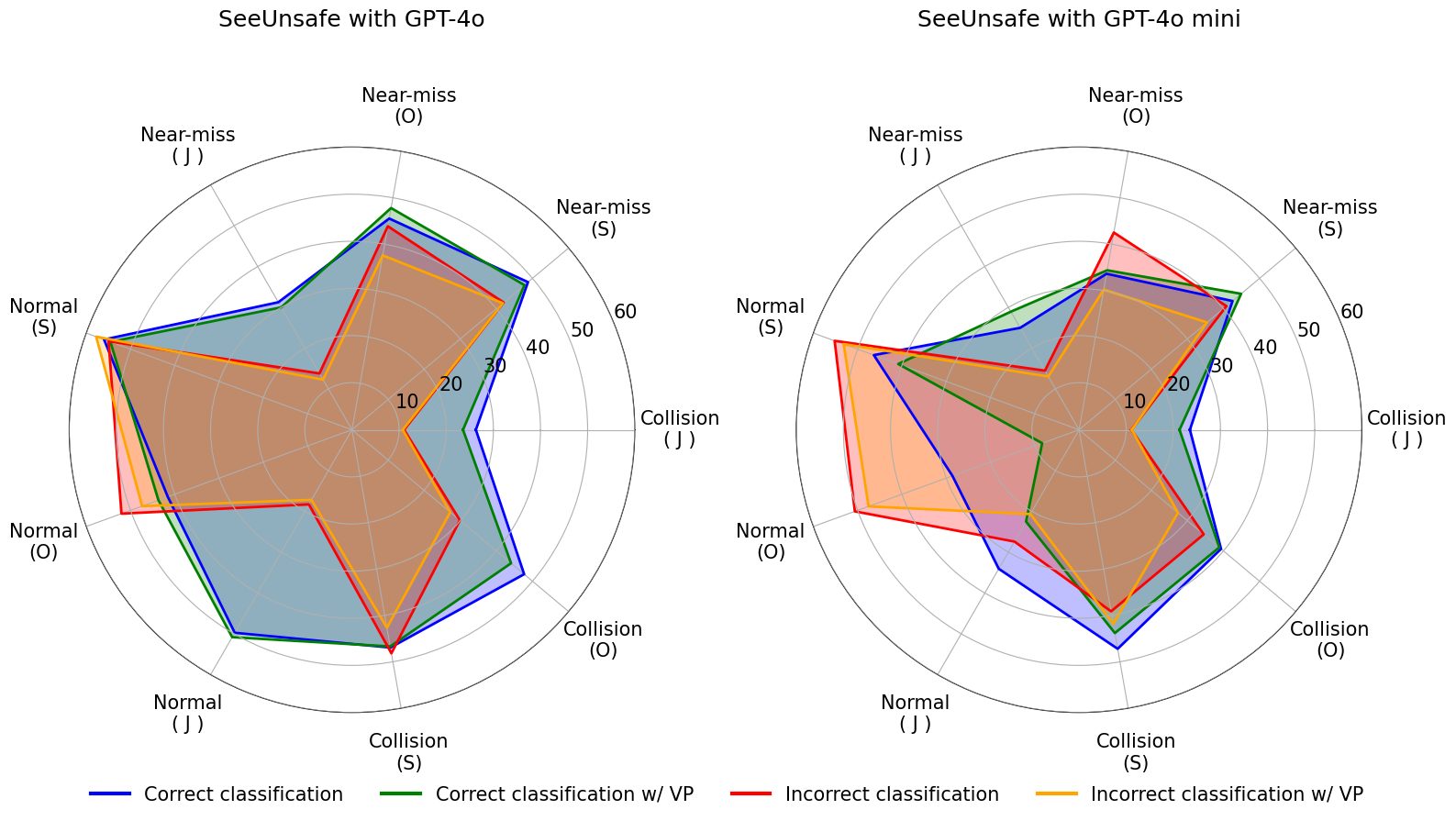}
	\caption{Evaluation of generated responses in both correct and incorrect classifications. We use S for scene context, O for object description, and J for justification.}
	\label{fig: experiment/radar_chart}
\end{figure}

We use the GPT-4o series within SeeUnsafe to explore the correlation between video classification accuracy and attribute-wise response alignment (against the ground truth). In Fig. \ref{fig: experiment/radar_chart}, we compare IMS values of each attribute (scene context, object description, and justification) obtained using GPT-4o and GPT-4o mini agents. Each radar plot includes four contours: correct classification, correct classification with visual prompts, incorrect classification, and incorrect classification with visual prompts. For GPT-4o, responses show better alignment when videos are correctly classified, and adding visual prompts minimally impacts the contour shape. In cases of incorrect classification, significant shrinkage is observed in the \texttt{justification} attribute for all classes, while \texttt{scene\_context} remains largely unaffected. This makes intuitive sense, as static backgrounds are easily identified regardless of object movements. The \texttt{object\_description} attribute shows inconsistent changes, decreasing for collision and near-miss videos but increasing for normal videos. For GPT-4o mini, the radar plot reveals greater shrinkage for correct classifications, especially when visual prompts are added, with the most noticeable decline in the normal class. Interestingly, IMS values for \texttt{scene\_context} and \texttt{object\_description} increase in some cases of incorrect classification. Upon examining failure cases, we found that misclassified normal videos often contain clear objects in the PoV, enabling the model to generate better-aligned responses despite the misclassification. These findings highlight the nuanced interplay between video classification and response alignment and imply that generating the response in a structured format can be a useful attempt to improve the interpretability of MLLM outputs.

\subsubsection{Performance under different lighting conditions}

Input quality plays a critical role in vision-based tasks, with factors such as image resolution, noise, lighting, and lens distortion affecting performance. This applies to our problem as well. In the context of traffic videos mostly sourced from surveillance cameras, resolution and lens distortion are typically fixed for a given camera, but lighting conditions vary with time and weather. To evaluate the robustness of existing MLLM agents under different lighting conditions, we analyze performance during daytime and nighttime scenarios. In detail, we find all 19 nighttime scenarios (6 collisions and 13 near-misses) in the WTS dataset and compare them against 19 randomly sampled daytime scenarios (from the remaining 126) over 200 repeats. As shown in Table \ref{tab: experiment/lighting_condition}, classification accuracy and response alignment are generally better during daytime than nighttime, as expected. Adding visual prompts improves daytime performance for collisions and near-misses by clarifying object boundaries and enhancing spatial relationships. However, in nighttime scenarios, visual prompts degrade performance, likely due to added noise. One explanation is that nighttime inputs already contain white noise, and the additional prompts further obscure the images, making object identification more challenging. Note that while we compared the 6 collision and 13 near-miss videos from nighttime with the same number of daytime videos (as shown in the first four rows under each lighting condition), the limited nighttime samples introduce bias. This results in higher separate evaluation scores for nighttime collisions and near-misses compared to daytime samples. The overall accuracy (as shown in the last two rows under each lighting condition) should be considered for a fair comparison.

\begin{table}
\caption{Performance under different lighting conditions. The rows in \colorbox{gray!30}{gray} show the results used for comparison.}
\label{tab: experiment/lighting_condition}
\resizebox{1\linewidth}{!}{
\begin{threeparttable}
\begin{tabular}{lllllllllll}
\toprule
\begin{tabular}[c]{@{}l@{}}Lighting\\ Condition\end{tabular} & Class  & Visual Prompt & Accuracy & Precision & F1-Score & BLEU\tnote{*} & ROUGE-L\tnote{*} & IMS\tnote{*} \\ \midrule
\multirow{6}{*}{Nighttime}    & \multirow{2}{*}{Near-miss (13)} & No &  15.38  & 100.00 & 26.67 &  36.28  &  36.80  &  31.62  \\
 &                            & Yes  &  15.38  &  100.00  &  26.67  &  35.97  &  35.71  &  
29.37  \\  
 & \multirow{2}{*}{Collision (6)} & No  &  100.00  &  100.00  &  100.00  &  32.02  &  34.01  &  38.55  \\
 &                            & Yes &  83.33  &  100.00  &  90.91  & 30.15  &  30.17  &  27.67  \\ 
 &  \multirow{2}{*}{ Overall (19)}  & No\cellcolor{gray!30}  &  42.11 \cellcolor{gray!30} &  80.26 \cellcolor{gray!30} &  35.47 \cellcolor{gray!30} &   34.93 \cellcolor{gray!30}  &   35.92 \cellcolor{gray!30} &  33.81 \cellcolor{gray!30}  \\ 
 &                            & Yes\cellcolor{gray!30}  & 36.84 \cellcolor{gray!30} & 56.89\cellcolor{gray!30}  & 32.89 \cellcolor{gray!30} & 34.14 \cellcolor{gray!30} & 33.96 \cellcolor{gray!30} &  28.84  \cellcolor{gray!30} \\ \hline
\multirow{6}{*}{Daytime} & \multirow{2}{*}{Near-miss (13)} & No &  10.46  &  82.80  &  18.26  &  38.78  &  38.24  &  34.05  \\
 &                            & Yes &  8.98  &  72.40  &  14.18  &  38.25  &  36.89  &  31.48  \\
 & \multirow{2}{*}{Collision (6)} & No &  90.43  &  100.00  &  94.49  &  35.38  &  37.18  &  39.24  \\
 &                            & Yes &  95.73  &  100.00  &  97.63  &  35.90  &  36.87  &  
37.78  \\
 & \multirow{2}{*}{Overall (19) } & No\cellcolor{gray!30}  &  68.18 \cellcolor{gray!30}  &  66.79 \cellcolor{gray!30}  &  64.10 \cellcolor{gray!30}  & 36.32  \cellcolor{gray!30}  &   37.49 \cellcolor{gray!30} &  37.79 \cellcolor{gray!30} 
 \\
 &                            & Yes \cellcolor{gray!30}  &  71.47 \cellcolor{gray!30}  &   66.50 \cellcolor{gray!30}  &   65.49\cellcolor{gray!30}   &   36.44 \cellcolor{gray!30}  &  36.82 \cellcolor{gray!30}  &  35.95 \cellcolor{gray!30}  \\
\bottomrule
\end{tabular}%
\begin{tablenotes}
  \item[*] Equally-weighted average over three components.
\end{tablenotes}
\end{threeparttable}
}
\end{table}

\subsection{Qualitative results}
\label{sec: experiment/qualitative_Result}
We provide qualitative examples for three video classes, one for each respectively. We start with a near-miss event in which a white sedan drives through an intersection and stops in time to avoid colliding with a male pedestrian. As shown in Table \ref{tab: experiment/example_near_miss}, the first clip is classified as normal because the vehicle and pedestrian are not present in the FoV. In the second clip, while the objects (\texttt{vehicle: 0} and \texttt{person: 1}) appear in the FoV, they are far apart, so it is also classified as normal. In the final clip, the vehicle approaches the pedestrian but stops at a safe distance, resulting in a near-miss classification. Using the severity-based aggregation method (Sec. \ref{sec: methodology/aggregation}), the video is classified as near-miss overall. The visual grounding results confirm that the involved objects are successfully highlighted with segmentation masks. Following the near-miss example, we present a collision scenario in Table \ref{tab: experiment/example_collision}. In this case, a white sedan approaches from behind and hits a pedestrian. In the first clip, only pedestrians are visible in the field of view (FoV), ending with being classified as normal. In the second clip, a white sedan appears from the right, but neither accelerates nor steers to avoid the pedestrian, resulting in contact. In the final clip, the sedan and the pedestrian remain in close contact and stop. Consequently, the last two clips are classified as collisions. Lastly, we present an example of a normal event in Table \ref{tab: experiment/example_normal}, which constitutes the majority of daily traffic camera footage. The scene includes several objects, such as pedestrians and vehicles. Since the textual prompt instructs the MLLM agent to generate descriptions for two objects, it automatically selects the two most visible and closest to the camera: the pedestrian on the crosswalk and the sedan making a right turn. These results indicate that SeeUnsafe can classify traffic events and identify road users involved in critical events. 

\begin{table}
\caption{Visualization of scenario $\text{20231006\_21\_CN5\_T1}$.}
\label{tab: experiment/example_near_miss}
\begin{threeparttable}
\begin{tabular}{p{0.31\textwidth} p{0.31\textwidth} p{0.31\textwidth}}
\toprule
Ground truth: Near-miss. & & \\ \midrule \includegraphics[width=0.31\textwidth]{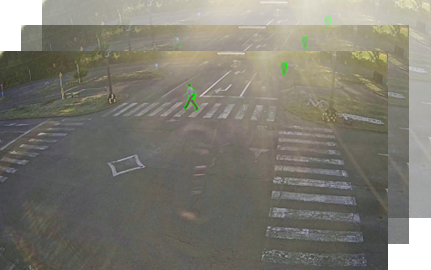} & \includegraphics[width=0.31\textwidth]{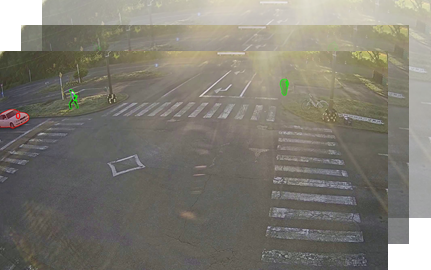} & \includegraphics[width=0.31\textwidth]{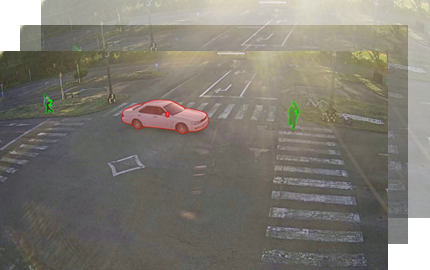}\\
\textbf{Event class}: Normal & \textbf{Event class}: Normal & \cellcolor{gray!30} \textbf{Event class}: Near-miss    \\
\textbf{Scene context}: Sunny day with dry road surface at a crosswalk near an intersection, under daytime lighting with light traffic. & \textbf{Scene context}: Sunny day with dry road surface in a curbside area near an intersection, under daytime lighting with moderate mixed traffic. & \cellcolor{gray!30} \textbf{Scene context}: Sunny day with dry road surface at a crosswalk near an intersection, under daytime lighting with moderate traffic.\\
\textbf{Object description}: Two pedestrians. & \textbf{Object description}: Two pedestrians and a white sedan & \cellcolor{gray!30} \textbf{Object description}: A pedestrian and a white sedan\\
\textbf{Justification}: The pedestrians are walking across the crosswalk at a steady pace without any sudden changes in speed or direction, and no vehicles are present in close proximity. & \textbf{Justification}: The pedestrians and the vehicle maintain their paths without any sudden changes in speed or direction, indicating no traffic anomaly or collision. & \cellcolor{gray!30} \textbf{Justification}: The pedestrian is crossing the road while the sedan approaches the crosswalk. The vehicle shows a significant deviation in its path to avoid the pedestrian, indicating a near-miss situation.\\ \\
\multicolumn{3}{l}{\textbf{Visual grounding}: car id: \{0\}, person id: \{1\}}  \\
\bottomrule
\end{tabular}%
\end{threeparttable}
\end{table}

\begin{table}[htbp]
\caption{Visualization of scenario $\text{20230728\_15\_SY18\_T2}$.}
\label{tab: experiment/example_collision}
\begin{threeparttable}
\begin{tabular}{p{0.31\textwidth} p{0.31\textwidth} p{0.31\textwidth}}
\toprule
Ground truth: Collision. & & \\ \midrule \includegraphics[width=0.32\textwidth]{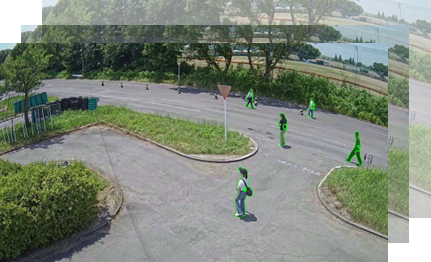} & \includegraphics[width=0.32\textwidth]{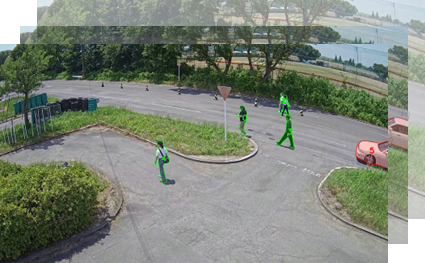} & \includegraphics[width=0.32\textwidth]{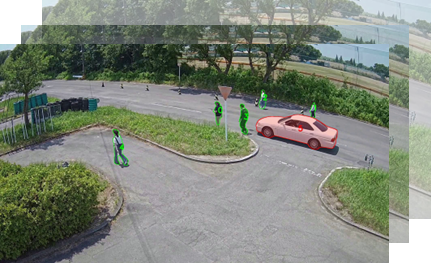}\\
\textbf{Event class}: Normal & \cellcolor{gray!30} \textbf{Event class}: Collision & \cellcolor{gray!30} \textbf{Event class}: Collision    \\
\textbf{Scene context}: Sunny day with dry road surface in a curbside area near an intersection, under daytime lighting with moderate mixed traffic. & \cellcolor{gray!30} \textbf{Scene context}: Sunny day with dry road surface in a curbside area near an intersection, under daytime lighting with moderate mixed traffic. & \cellcolor{gray!30} \textbf{Scene context}: Sunny day with dry road surface in a curbside area near an intersection, under daytime lighting with moderate mixed traffic.\\
\textbf{Object description}: Multiple pedestrians. & \cellcolor{gray!30} \textbf{Object description}: A white sedan and a male pedestrian. & \cellcolor{gray!30} \textbf{Object description}: A pedestrian and a white sedan\\
\textbf{Justification}: The pedestrians maintain a consistent walking pace and direction without any sudden changes or interactions that suggest a near-miss or collision. & \cellcolor{gray!30} \textbf{Justification}: The pedestrian's path intersects with the vehicle's trajectory, and the vehicle does not deviate or stop, resulting in a collision with the pedestrian. & \cellcolor{gray!30} \textbf{Justification}: The pedestrian is in direct contact with the sedan, indicating a collision. The vehicle's position suggests it failed to stop in time, resulting in the impact.
\\ \\
\multicolumn{3}{l}{\textbf{Visual grounding}: car id: \{5\}, person id: \{3\}}  \\
\bottomrule
\end{tabular}%
\end{threeparttable}
\end{table}

\begin{table}[htbp]
\caption{Visualization of scenario $\text{20231013\_101824\_normal\_192.168.0.11\_1\_event\_2}$.}
\label{tab: experiment/example_normal}
\begin{threeparttable}
\begin{tabular}{p{0.31\textwidth} p{0.31\textwidth} p{0.31\textwidth}}
\toprule
Ground truth: Normal. & & \\ \midrule \includegraphics[width=0.32\textwidth]{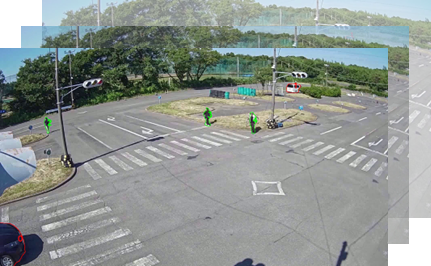} & \includegraphics[width=0.32\textwidth]{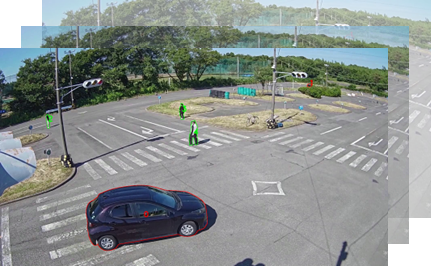} & \includegraphics[width=0.32\textwidth]{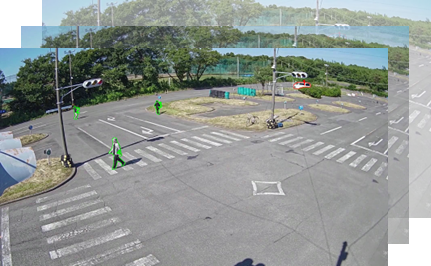}\\
\cellcolor{gray!30} \textbf{Event class}: Normal. & \cellcolor{gray!30} \textbf{Event class}: Normal. & \cellcolor{gray!30} \textbf{Event class}: Normal.    \\
\cellcolor{gray!30} \textbf{Scene context}: Sunny day with dry road surface in a curbside area near an intersection, under daytime lighting with moderate mixed traffic. & \cellcolor{gray!30} \textbf{Scene context}: Sunny day with dry road surface in a parking lot area near an intersection, under daytime lighting with light traffic. & \cellcolor{gray!30} \textbf{Scene context}: Sunny day with dry road surface in a parking lot area near an intersection, under daytime lighting with light traffic.\\
\cellcolor{gray!30} \textbf{Object description}: A pedestrian and a black sedan. & \cellcolor{gray!30} \textbf{Object description}: A pedestrian and a black sedan. & \cellcolor{gray!30} \textbf{Object description}: A pedestrian and a black sedan.\\
\cellcolor{gray!30} \textbf{Justification}: There are no sudden deviations in speed, direction, or proximity between the pedestrian and the vehicle. The pedestrian and vehicle maintain their paths without any indication of a near-miss or collision. & \cellcolor{gray!30} \textbf{Justification}: The pedestrian and the vehicle maintain their paths without any sudden changes in speed or direction, indicating no traffic anomaly. & \cellcolor{gray!30} \textbf{Justification}: The pedestrian and the vehicle maintain their paths without any sudden changes in speed or direction, indicating no traffic anomaly or risk of collision.
\\ \\
\multicolumn{3}{l}{\textbf{Visual grounding}: N/A}  \\
\bottomrule
\end{tabular}%
\end{threeparttable}
\end{table}
\section{Discussion}
\label{sec: discussion}
\subsection{From traffic camera to dash camera}
\label{sec:discussion/overhead_vehicle}

\begin{table}
\caption{Multi-view video classification alignment of overhead surveillance camera and in-vehicle dash camera.}
\label{tab: discussion/overhead_vehicle}
\begin{threeparttable}
\begin{tabular}{p{0.31\textwidth} p{0.31\textwidth} p{0.31\textwidth}}
\toprule
Ground truth: Collision. &  & \\ \midrule \includegraphics[width=0.32\textwidth]{figure/collision_1.png} & \includegraphics[width=0.32\textwidth]{figure/collision_2.png} & \includegraphics[width=0.32\textwidth]{figure/collision_3.png}\\
\textbf{Event class}: Normal & \cellcolor{gray!30} \textbf{Event class}: Collision & \cellcolor{gray!30} \textbf{Event class}: Collision    \\
\textbf{Scene context}: Sunny day with dry road surface in a curbside area near an intersection, under daytime lighting with moderate mixed traffic. & \cellcolor{gray!30} \textbf{Scene context}: Sunny day with dry road surface in a curbside area near an intersection, under daytime lighting with moderate mixed traffic. & \cellcolor{gray!30} \textbf{Scene context}: Sunny day with dry road surface in a curbside area near an intersection, under daytime lighting with moderate mixed traffic.\\
\textbf{Object description}: Multiple pedestrians. & \cellcolor{gray!30} \textbf{Object description}: A white sedan and a male pedestrian. & \cellcolor{gray!30} \textbf{Object description}: A pedestrian and a white sedan\\
\textbf{Justification}: The pedestrians maintain a consistent walking pace and direction without any sudden changes or interactions that suggest a near-miss or collision. & \cellcolor{gray!30} \textbf{Justification}: The pedestrian's path intersects with the vehicle's trajectory, and the vehicle does not deviate or stop, resulting in a collision with the pedestrian. & \cellcolor{gray!30} \textbf{Justification}: The pedestrian is in direct contact with the sedan, indicating a collision. The vehicle's position suggests it failed to stop in time, resulting in the impact.
\\ \\
\multicolumn{3}{l}{\textbf{Visual grounding}: car id: \{5\}, person id: \{3\}}  \\
\midrule \includegraphics[width=0.32\textwidth]{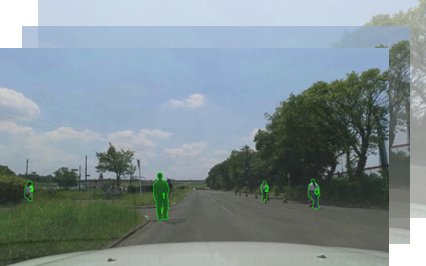} & \includegraphics[width=0.32\textwidth]{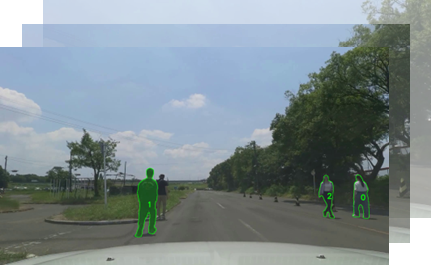} & \includegraphics[width=0.32\textwidth]{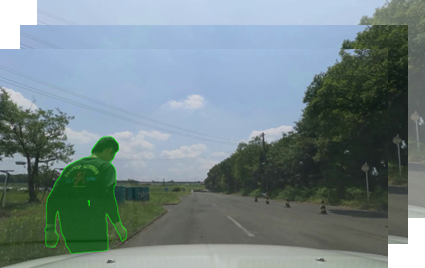}\\
\textbf{Event class}: Normal & \cellcolor{gray!30} \textbf{Event class}: Collision & \cellcolor{gray!30} \textbf{Event class}: Collision    \\
\textbf{Scene context}: A sunny day with clear skies and light traffic on a rural road with trees lining the sides. & \cellcolor{gray!30} \textbf{Scene context}: A sunny day with clear skies on a rural road with some traffic cones on the side. & \cellcolor{gray!30} \textbf{Scene context}: A sunny day with clear skies on a rural road with traffic cones lining the side.\\
\textbf{Object description}: Pedestrians walking on the side of the road. & \cellcolor{gray!30} \textbf{Object description}: A pedestrian wearing a black shirt. & \cellcolor{gray!30} \textbf{Object description}: A pedestrian wearing a dark shirt.\\
\textbf{Justification}: The pedestrians maintain a consistent path on the side of the road, and the ego vehicle continues on its course without any sudden changes in speed or direction, indicating a normal driving situation. & \cellcolor{gray!30} \textbf{Justification}: The pedestrian is initially walking along the road and then suddenly moves into the path of the ego vehicle. The ego vehicle does not appear to take evasive action, resulting in a collision with the pedestrian. & \cellcolor{gray!30} \textbf{Justification}: The pedestrian in the dark shirt is directly in the path of the ego vehicle, indicating a failure to avoid the pedestrian, resulting in a collision. The ego vehicle does not show signs of swerving or braking in time to prevent the impact.
\\ \\
\multicolumn{3}{l}{\textbf{Visual grounding}: person id: \{1\}}  \\
\bottomrule
\end{tabular}%
\end{threeparttable}
\end{table}

Although this work focuses on helping traffic managers process large volumes of traffic camera footage, the framework can also be applied in vehicles equipped with dash cameras. Leveraging the Toyota WTS dataset, which includes in-vehicle footage, we present a case study to demonstrate the potential for using dash cameras to provide real-time surrogate safety measurements from the driver’s perspective. As shown in Table \ref{tab: discussion/overhead_vehicle}, we compare video classification and visual grounding results for the same events captured from two perspectives: an overhead surveillance camera and an in-vehicle dash camera. The images from both perspectives are synchronized, yielding the same classification for each clip and correctly classifying the entire event. Interestingly, the in-vehicle footage, with its smaller field of view (FoV), generates more detailed responses compared to the surveillance footage. This highlights the effectiveness of in-vehicle dash camera footage for evaluating driving risks and its potential for real-time risk assessment. This example suggests the potential for collaborative video understanding in multi-view settings, useful when object interactions are occluded in surveillance but visible from in-vehicle views.

\subsection{Limitations}
Our work is among the first to explore using MLLMs for risk-aware video classification and visual grounding to identify road users in critical events. However, as MLLMs are still in their early stages of solving complex vision tasks, this work has suffered from some limitations. Firstly, our work defines critical traffic events as collisions and near-misses, but identifying near-miss events is challenging. As shown in Fig. \ref{fig: experiment/confusion_matrix} and Fig. \ref{fig: experiment/confusion_matrix_main}, near-misses are often confused with collisions, especially under occlusion or limited visibility from a single camera view. This limitation is common in vision-based tasks, where a single perspective provides only partial observations. Multi-view video footage, such as combining surveillance and in-vehicle views (Table \ref{tab: discussion/overhead_vehicle}), can mitigate these issues by offering complementary perspectives. Additionally, failures in visual grounding tasks often stem from inaccurate detection and tracking, highlighting the need for improved detection and tracking models to enhance the quality of visual prompts. From a framework standpoint, the current splitting and aggregation functions are intuitive and could benefit from advanced techniques. For example, key frame selection methods could filter irrelevant footage, while robust aggregation methods could address information uncertainty. Finally, MLLM performance could be enhanced by fine-tuning with customized datasets and incorporating human feedback tailored to traffic accident analysis tasks. Addressing these limitations would significantly improve the robustness and reliability of the SeeUnsafe framework in real-world applications.

\subsection{Broader impacts}

The SeeUnsafe framework leverages Multimodal Large Language Models (MLLMs) to tackle the challenge of processing vast amounts of unprocessed traffic video data, which often goes overlooked due to limited processing capabilities. By enabling efficient analysis, SeeUnsafe helps uncover previously unidentified accidents and near-misses, enhancing situational awareness and contributing to improved road safety. Its structured outputs can be stored in a database to facilitate content-based video retrieval, further supporting traffic management and research. Beyond surveillance footage, the framework’s adaptability allows deployment in vehicles as a real-time surrogate safety measure, providing instant insights into critical traffic scenarios. For vehicles equipped with dash cameras, SeeUnsafe offers a deeper understanding of driver behavior in high-risk situations, generating valuable data to improve driver training, refine traffic policies, and design safer road systems. By streamlining video analysis and situational awareness, SeeUnsafe has the potential to revolutionize traffic management and safety research, paving the way for a safer, more efficient, intelligent transportation system.

\section{Conclusions}
We present SeeUnsafe, a novel framework for accident-aware video classification and visual grounding that leverages Multimodal Large Language Models as the central agent for orchestrating these tasks. Empirical evaluations show that SeeUnsafe excels in traffic video understanding, offering a user-friendly, zero-shot, and scalable solution to process vast amounts of traffic footage without the need for extensive technical adaptation. SeeUnsafe is not only able to classify videos with 76.31\% accuracy but is also capable of performing visual grounding at a success rate of 51.47\%, which achieves the best performance compared to vanilla GPT-4o and other state-of-the-art models. This work marks an important step in leveraging MLLMs to transform unprocessed video data into actionable, safety-related insights for a range of applications. Looking ahead, several directions for improvement remain. The current uniform video splitting and severity-based aggregation methods, while effective, can benefit from advanced keyframe sampling and uncertainty-aware aggregation to handle varying accident rates and reduce output noise. Single-view input is prone to occlusion and can be enriched with multi-view footage from traffic or in-vehicle cameras. Lastly, the computational demands of MLLM-based methods highlight the need for lightweight models suitable for resource-constrained settings. Despite these limitations, SeeUnsafe establishes a promising baseline for integrating MLLMs into traffic monitoring systems, paving the way for safer and more efficient traffic management.

\section*{Acknowledgment}
The contents of this paper reflect the views of the authors, who are responsible for the facts and the accuracy of the information presented herein. The work is funded, partially or entirely, by NSF Grants 2426993 and 2322242, and the C2SMARTER Center under Grant Number 69A3552348326 from the U.S. Department of Transportation's University Transportation Centers Program. However, the U.S. Government assumes no liability for the contents or use thereof.  The work is also supported by the NYU IT High Performance Computing resources, services, and staff expertise.

\bibliographystyle{unsrtnat}
\bibliography{references}  

\appendix
\section{Appendix}
\label{section: appendix}

\subsection{Complete textual prompts}
\label{appendix: prompt}

The complete prompts used for the experiments are provided as follows. 
\begin{lstlisting}[language=Python, caption=Overhead Classification Prompt]
## prompt for overhead video classification
[system prompt]
- You are a traffic accident inspector. You need to determine which class ((*@\textcolor{black!30!green}{normal}@*), (*@\textcolor{black!10!orange}{near-miss}@*), or (*@\textcolor{black!10!red}{collision}@*) among road users) the provided videos are. Notice that some but not all of the pedestrians are marked with green contour and some but not all of the cars are marked with red contour. Use that to help with your observation.
- The normal class refers to no sudden deviations in speed, direction, or proximity between cars or between cars and pedestrians, or no objects are detected in the videos.
- The near-miss class depicts situations where vehicles or pedestrians come extremely close to colliding but ultimately avoid impact. These events often involve abrupt changes in speed, direction, or proximity, indicating a high risk of an accident that was narrowly avoided.
- The collision class depicts events where vehicles, pedestrians, or objects come into direct contact, resulting in an impact. These incidents often involve significant changes in speed or direction due to the collision, and can lead to visible damage or injury.
- We use 0 for near-miss, 1 for collision, and 2 for normal to represent the video class.
[user prompt]
- Three images are arranged in chronological order, depicting a traffic event. Please choose one from (*@\textcolor{black!30!green}{normal}@*), (*@\textcolor{black!10!orange}{near-miss}@*), and (*@\textcolor{black!10!red}{collision}@*) that can best describe the given video.
- If so, what types of road users are involved, and in what context do these traffic anomalies occur?
- Respond in the format of the following example without any additional information:
- Video Class: type of integer indicating the video class. For example: 0 for near-miss, 1 for collision, and 2 for normal.
- Object Description: type of string describing the appearance of road users. For example: The involved road users are a middle-aged male pedestrian and a white sedan
- Scene Context: type of string describing the scene environment. For example: Sunny day with dry road surface in a curbside area near an intersection, under daytime lighting with moderate mixed traffic.
- Justification: type of string explaining the reason why a traffic anomaly occurs or why a traffic anomaly doesn't exist. For example: The pedestrian's pose changes from walking to a sudden stop, indicating an unexpected reaction to the vehicle. The vehicle shows a significant deviation from its original trajectory, indicating that it misjudged the pedestrian's path. This heavy deviation led to the vehicle failing to avoid the pedestrian, resulting in a collision.          
\end{lstlisting}

\begin{lstlisting}[language=Python, caption=Vehicle View Classification Prompt]
## prompt for vehicle view video classification
[system prompt]
- You are a traffic accident inspector analyzing videos from an in-vehicle camera. Your task is to classify the video into one of three categories: (*@\textcolor{black!30!green}{normal}@*), (*@\textcolor{black!10!orange}{near-miss}@*), or (*@\textcolor{black!10!red}{collision}@*). The video shows the scene from the driver's perspective, and you should consider interactions between the ego vehicle and other road users visible ahead, such as pedestrians, vehicles, or obstacles.
- The normal class refers to situations where there are no sudden changes in speed or direction, and the ego vehicle maintains a safe distance from other road users or obstacles.
- The near-miss class refers to situations where the ego vehicle comes extremely close to colliding with another vehicle, pedestrian, or object but avoids impact. These situations often involve rapid changes in speed or direction, such as swerving or braking.
- The collision class refers to situations where the ego vehicle directly impacts another vehicle, pedestrian, or object. These events often involve significant changes in speed or direction due to the collision, with visible damage or injury.
- We use 0 for near-miss, 1 for collision, and 2 for normal to represent the video class.
Note: In the object details, only describe other road users (pedestrians, vehicles, or obstacles) other than the ego vehicle. The ego vehicle is assumed to be the vehicle with the camera.
- In the justification, describe the behaviors of both the ego vehicle and the involved road users. Focus on how the ego vehicle reacts to the behavior of other road users, such as braking, swerving, accelerating, or maintaining course, and how it influences or is influenced by the other objects.
- You will receive a sequence of 3 images as well as the corresponding time step as input. They are arranged in chronological order, representing events that are observed sequentially. The time step is 0.1 seconds. Based on this, determine the video class and answer questions.
[user prompt]
- Three images are arranged in chronological order, depicting a traffic event from the driver's perspective. Please choose one from (*@\textcolor{black!30!green}{normal}@*), (*@\textcolor{black!10!orange}{near-miss}@*), or (*@\textcolor{black!10!red}{collision}@*) that can best describe the given video.
- If so, what types of road users are involved, and in what context do these traffic anomalies occur?
- Respond in the format of the following example without any additional information:",
- Video Class: type of integer indicating the video class. For example: 0 for near-miss, 1 for collision, and 2 for normal.
- Object Description: type of string describing the appearance of the road users other than the ego vehicle. For example: The involved road user is a pedestrian wearing a red helmet.
- Scene Context: type of string describing the scene environment. For example: A sunny day with clear skies and light traffic on a residential street with parked cars lining the road.
- Justification: type of string explaining the reason why a traffic anomaly occurs or why a traffic anomaly doesn't exist. For example: The pedestrian's pose changes from walking to a sudden stop, indicating an unexpected reaction to the ego vehicle. The ego vehicle shows a significant deviation from its original trajectory, indicating that it misjudged the pedestrian's path. This heavy deviation led to the vehicle failing to avoid the pedestrian, resulting in a collision.
\end{lstlisting}

\begin{lstlisting}[language=Python, caption=Visual Grounding Prompt]
## prompt for visual grounding
[user prompt]
- The following video contains near-miss or collision traffic incidents. The objects involved in the incident have been labeled with numerical IDs. Please identify the IDs of the involved objects from the video. Specifically, pedestrians are labeled in green, and cars are labeled in red.
Note that there exists and only exixts one pedestrian and car involved. So you should only answer with one ID for pedestrian and one ID for car.
Response format example:
(*@\textcolor{black!30!green}{Pedestrian ID: 3}@*)
(*@\textcolor{black!10!red}{Car ID: 5}@*)         
\end{lstlisting}

\begin{lstlisting}[language=Python, caption=Detailed Information Matching Prompt, label=lst:detailed_visual_perception_prompt]
## Prompt MLLM to evaluate the matching quality
- You are tasked with evaluating the overall matching quality between the ground truth and predicted descriptions for three components: scene context, object description, and justification.
- Please carefully consider the latent correlations among these components and provide individual scores for each.
- 1. (*@\bf{Scene\_Context:}@*) Focus on the factors such as weather, lighting conditions, and road topology (i.e. intersection). These are representative factors to categorize the analyzed video. For missing content in the predicted descriptions, consider them as wrong.
- 2. (*@\bf{Object\_Description:}@*) Focus on the factors including number of objects, the class of objects, and appearance information. For missing content in the predicted descriptions, consider them as wrong.
- 3. (*@\bf{Justification:}@*)  This is the most important component. Be strict and rigorous when evaluating this component. Pay attention to descriptions regarding the dynamics, movement, and relative positions of the involved objects. If you think the predicted description misses important contents regarding the ground truth or incorrect, assign a low score.
- Please provide a score from 0 to 100 for each component.
- 100: Perfect alignment with no missing or incorrect information.
- 80(*@-@*)99: Very good alignment with only minor discrepancies.
- 60(*@-@*)79: Good alignment but with some significant missing or incorrect details.
- 40(*@-@*)59: Moderate alignment with noticeable errors or omissions.
- 20(*@-@*)39: Poor alignment with many incorrect or missing details.
- 0(*@-@*)19: Completely incorrect or irrelevant prediction.

- Ground Truth:
    - Scene Context: {gt_scene_context}
    - Object Description: {gt_object_detail}
    - Justification: {gt_justification}

- Predicted:
    - Scene Context: {pred_scene_context}
    - Object Description: {pred_object_detail}
    - Justification: {pred_justification}

- Respond in the format of the following example without any additional information:
- (*@\bf{Scene\_Context\_Score:}@*) type of int number indicating the matching quality of the scene context. For example: 85
- (*@\bf{Object\_Description\_Score:}@*) type of int number indicating the matching quality of the object detail. For example: 75
- (*@\bf{Justification\_Score:}@*) type of int number indicating the matching quality of the justification. For example: 90
\end{lstlisting}

\subsection{Dataset statistics}
\label{appendix: dataset}
This table shows the number (per class) and duration of videos used in the work. 
\begin{table}[h]
\centering
\caption{Statistics of the dataset used for video classification.}
\label{tab: appendix/dataset statistics}
\begin{threeparttable}
\begin{tabular}{llll}
\toprule
          & Mean (sec) & Std (sec) & Number \\ \midrule
Near-miss & 77.38      & 28.78     & 39     \\
Collision & 85.91      & 27.44     & 106    \\
Normal    & 29.90      & 23.11     & 104    \\
\bottomrule
\end{tabular}%
\end{threeparttable}
\end{table}

\subsection{Evaluation of IMS}
\label{appendix: evaluation ims}
We provide additional visualizations for comparing our proposed IMS and existing NLP metrics, BLEU and ROUGE-L, on testing different MLLM agents. It can be seen that there is a consistent positive correlation between the three attributes when using IMS. Additionally, when using GPT-4 for calculating IMS, points are more clustered in the upper right corner with brighter colors, indicating better response quality perceived by the MLLM evaluator. 
\begin{figure}[htbp]
	\centering
	\includegraphics[width=0.7\textwidth]{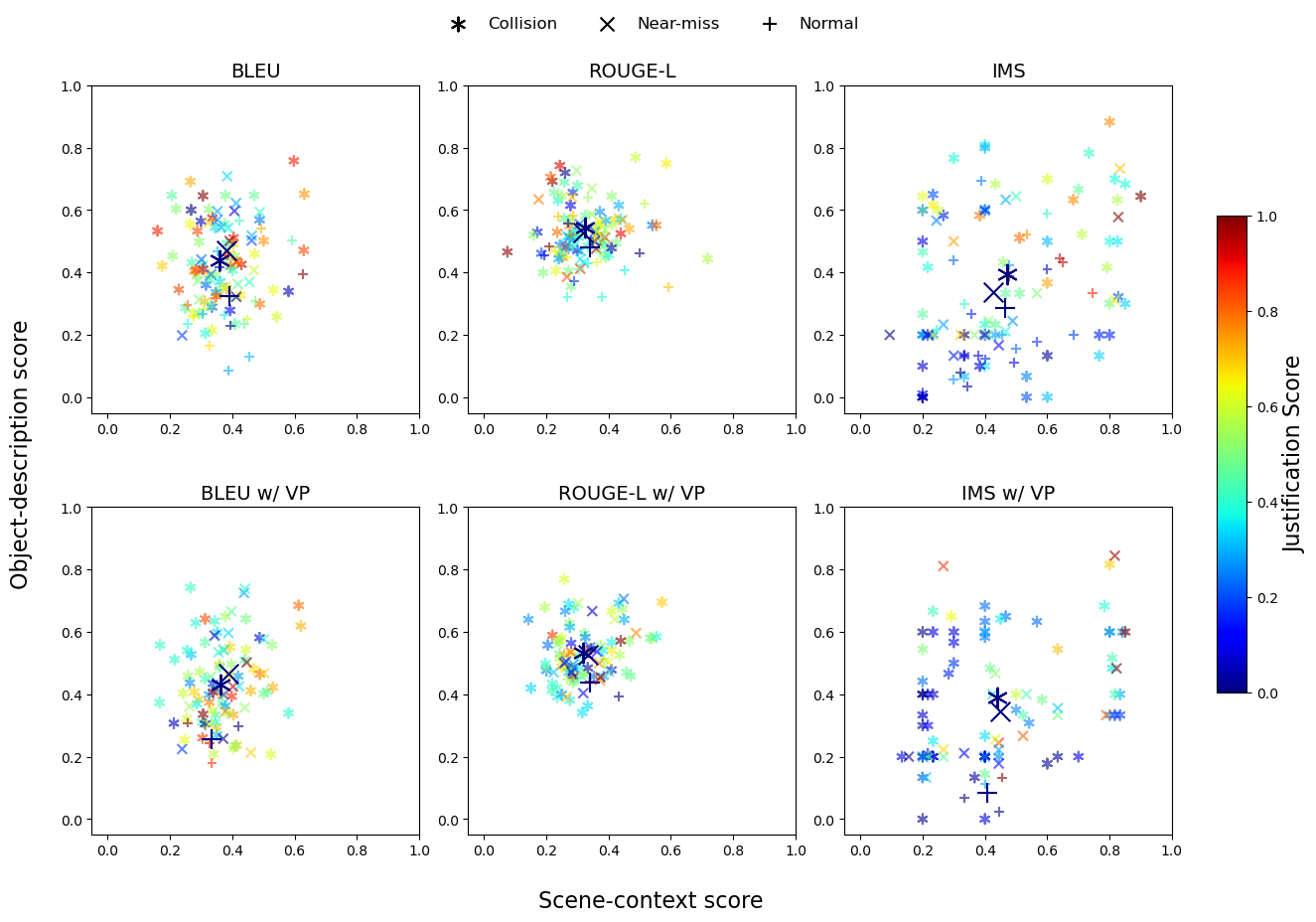}
	\caption{Qualitative evaluation of BLEU, ROUGE-L, and IMS for responses of SeeUsafe with a GPT-4o mini agent.}
	\label{fig: appendix/nlp_gpt4o-mini}
\end{figure}

\begin{figure}[htbp]
	\centering
	\includegraphics[width=0.7\textwidth]{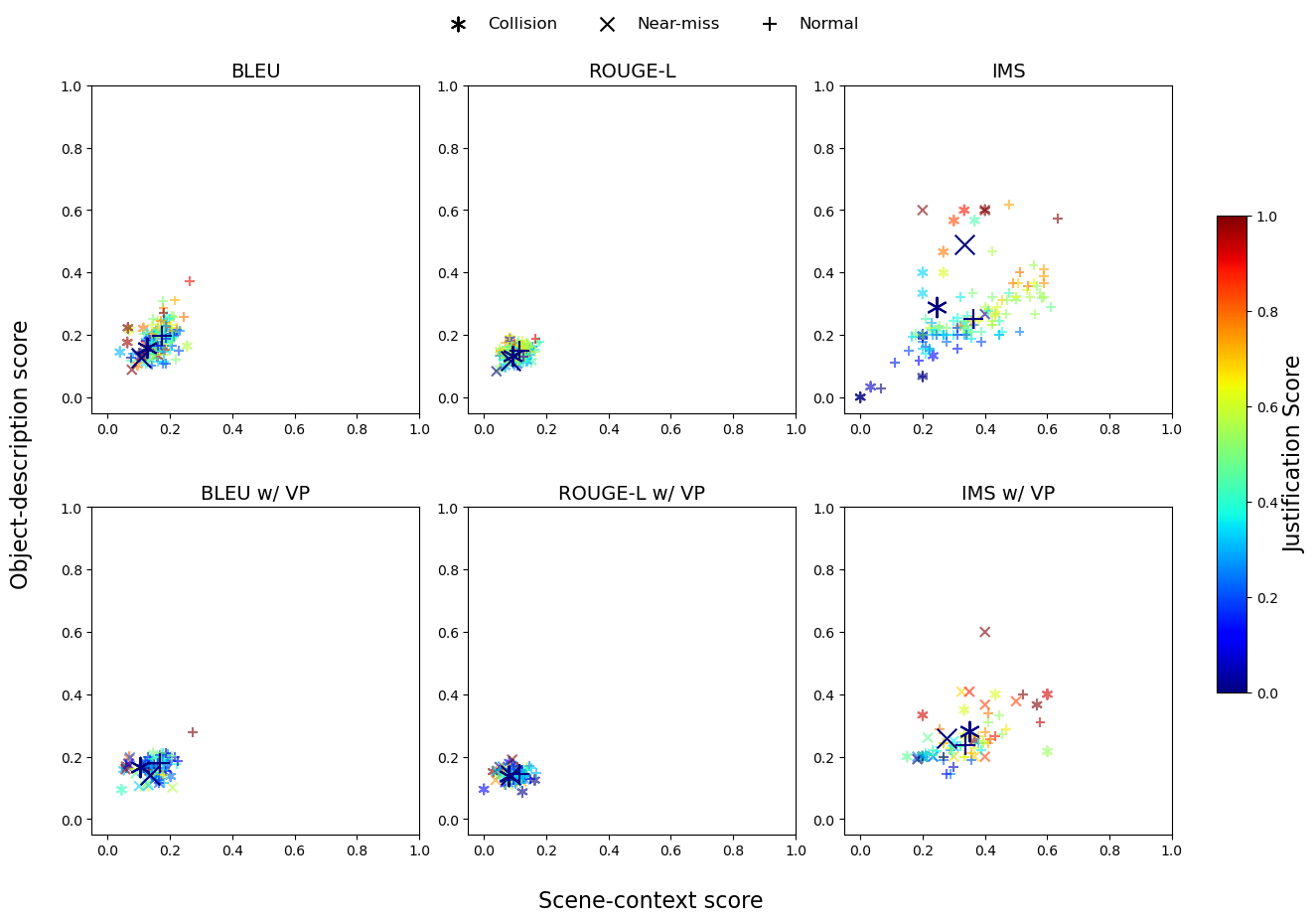}
	\caption{Qualitative evaluation of BLEU, ROUGE-L, and IMS for responses of SeeUsafe with an $\text{LLaVA-NeXT}_{interleave}$ agent.}
	\label{fig: appendix/nlp_gpt4o-mini}
\end{figure}

\begin{figure}[htbp]
	\centering
	\includegraphics[width=0.7\textwidth]{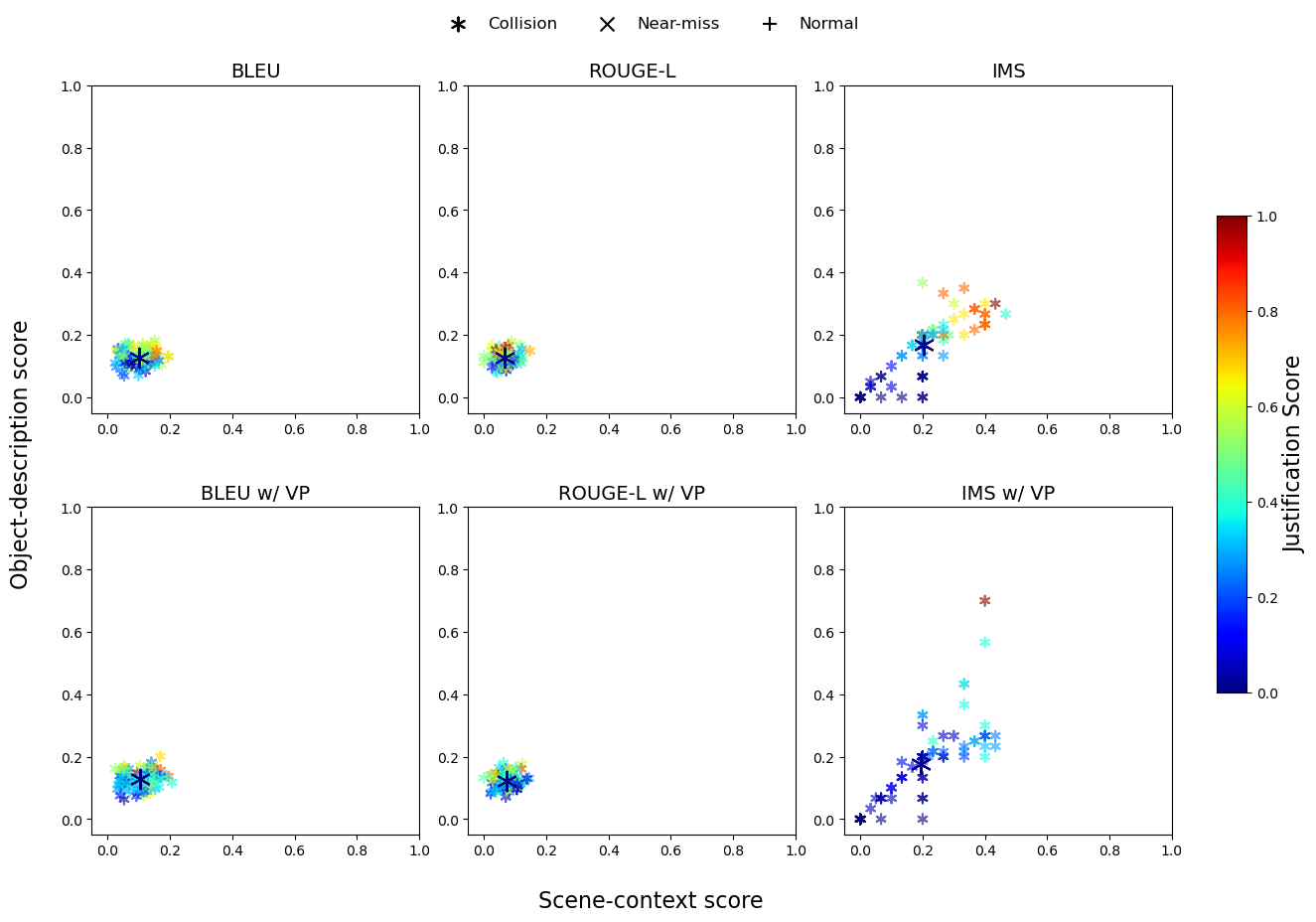}
	\caption{Qualitative evaluation of BLEU, ROUGE-L, and IMS for responses of SeeUsafe with an $\text{LLaVA-NeXT}_{video}$ agent.}
	\label{fig: appendix/nlp_gpt4o-mini}
\end{figure}

\begin{figure}[t]
	\centering
	\includegraphics[width=0.7\textwidth]{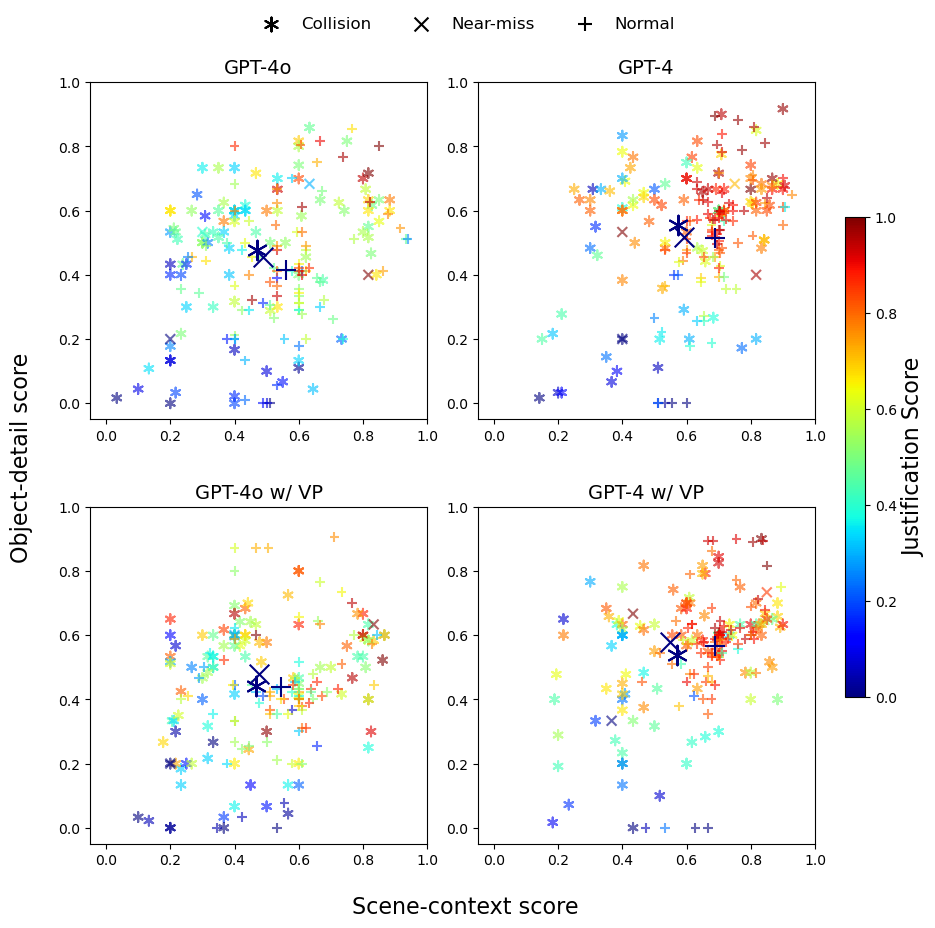}
	\caption{Qualitative comparison of IMS computed by GPT-4o and GPT-4 for responses of SeeUsafe with a GPT-4o agent.}
	\label{fig: appendix/ims_agent_comparision}
\end{figure}






\end{document}